\documentclass[conference]{IEEEtran}
\usepackage{times}

\let\labelindent\relax
\usepackage[shortlabels]{enumitem}
\usepackage[numbers]{natbib}
\usepackage{multicol}
\usepackage[bookmarks=true]{hyperref}

\usepackage{algorithm}
\usepackage{algorithmic}

\usepackage{amssymb}
\usepackage{amsmath}
\usepackage{amssymb}
\usepackage{mathtools}
\usepackage{mathrsfs}
\usepackage{commath}
\usepackage{multirow}
\usepackage[normalem]{ulem}

\usepackage{tabularx}
\usepackage{booktabs}
\usepackage{array}

\usepackage{comment}
\usepackage{wrapfig}

\usepackage{cuted}

\usepackage{xcolor}

\usepackage{tikz}
\usetikzlibrary{automata, positioning, arrows.meta, calc}

\usepackage{amsthm}
\newtheorem{theorem}{Theorem}

\newtheorem{corollary}{Corollary}

\theoremstyle{definition}

\newtheorem{assumption}{Assumption}

\def\equationautorefname~#1\null{(#1)\null}
\def\sectionautorefname~#1\null{Section #1\null}
\def\subsectionautorefname~#1\null{Section #1\null}
\def\algorithmautorefname~#1\null{Algorithm #1\null}
\def\theoremautorefname~#1\null{Theorem #1\null}
\def\lemmaautorefname~#1\null{Lemma #1\null}
\def\corollaryautorefname~#1\null{Corollary #1\null}
\def\remarkautorefname~#1\null{Remark #1\null}
\def\definitionautorefname~#1\null{Definition #1\null}
\def\assumptionautorefname~#1\null{Assumption #1\null}
\begin{document}

\title{EigenSafe: A Spectral Framework for Learning-Based Probabilistic Safety Assessment}

% You will get a Paper-ID when submitting a pdf file to the conference system
% \author{Author Names Omitted for Anonymous Review. Paper-ID [XXX]}

% \author{\authorblockN{Michael Shell}
% \authorblockA{School of Electrical and\\Computer Engineering\\
% Georgia Institute of Technology\\
% Atlanta, Georgia 30332--0250\\
% Email: mshell@ece.gatech.edu}
% \and
% \authorblockN{Homer Simpson}
% \authorblockA{Twentieth Century Fox\\
% Email: homer@thesimpsons.com}
% \and
% \authorblockN{James Kirk\\ and Montgomery Scott}
% \authorblockA{Starfleet Academy\\
% San Francisco, California 96678-2391\\
% Telephone: (800) 555--1212\\

% avoiding spaces at the end of the author lines is not a problem with
% conference papers because we don't use \thanks or \IEEEmembership


\author{\authorblockN{Inkyu Jang\authorrefmark{1}\authorrefmark{4},
Jonghae Park\authorrefmark{1}\authorrefmark{4},
Sihyun Cho\authorrefmark{2}, 
Chams E. Mballo\authorrefmark{3}, 
Claire J. Tomlin\authorrefmark{3} and
H. Jin Kim\authorrefmark{4}}
\authorblockA{\authorrefmark{4}Department of Aerospace Engineering,
Seoul National University,
Seoul, Korea %Email: mshell@ece.gatech.edu
}
\authorblockA{\authorrefmark{2}Interdisciplinary Program in Artificial Intelligence, Seoul National University, Seoul, Korea}
\authorblockA{\authorrefmark{3}University of California, Berkeley, CA, USA}
% % Starfleet Academy, San Francisco, California 96678-2391\\
% % Telephone: (800) 555--1212, Fax: (888) 555--1212}
\authorblockA{\authorrefmark{1}Equal Contribution}
}

% \IEEEpeerreviewmaketitle

\maketitle

\begin{abstract}
We present EigenSafe, an operator-theoretic framework for safety assessment of learning-enabled stochastic systems. In many robotic applications, the dynamics are inherently stochastic due to factors such as sensing noise and environmental disturbances, and  it is challenging for conventional methods such as Hamilton-Jacobi reachability and control barrier functions to provide a well-calibrated safety critic that is tied to the actual safety probability. We derive a linear operator that governs the dynamic programming principle for safety probability, and find that its dominant eigenpair provides critical safety information for both individual state-action pairs and the overall closed-loop system. The proposed framework learns this dominant eigenpair, which can be used to either inform or constrain policy updates. We demonstrate that the learned eigenpair effectively facilitates safe reinforcement learning. Further, we validate its applicability in enhancing the safety of learned policies from imitation learning through robot manipulation experiments using a UR3 robotic arm in a food preparation task.

Project Webpage: \textcolor{magenta}{\texttt{\url{https://eigen-safe.github.io/}}}
\end{abstract}

\section{Introduction}
\subsection{Motivation}

Ensuring safe operation is a fundamental requirement for the deployment of autonomous robotic systems, yet achieving it is challenging due to the inherent stochasticity of real-world operations. Deployed robots face unpredictable environmental disturbances, sensor noise, and the uncertainties introduced by learning-enabled components. In these stochastic settings, the traditional \textit{binary} view of safety, where a state is strictly classified as either safe or unsafe, becomes unrealistic. Instead, safety is quantified according to the probability of not experiencing any failure over a given time horizon.

Therefore, it is essential to accurately estimate the closed-loop safety probability and understand how it evolves over long time horizons. Achieving this requires a safety critic that can evaluate this probability. Prior approaches based on optimal control, e.g.,  Hamilton-Jacobi (HJ) reachability analysis \cite{bansal2017hamilton, ganai2024hamilton} and learning-based control barrier functions (CBFs) \cite{ames2019control, so2024train}, have successfully demonstrated their capabilities as learning-based safety critics. However, they typically learn surrogate values, such as value functions or barrier scores, which do not inherently correspond to the actual safety probability. Moreover, because these critics are not explicitly calibrated to the closed-loop safety probability, their outputs can be inconsistent or overly conservative, leading to overly cautious policies that unnecessarily sacrifice performance.

These issues can be resolved by using a safety critic that is \textit{well-calibrated} to the evolution of the true closed-loop safety probability over long time horizons. 
A critic that serves as a consistent proxy for the actual probability would enable a precise trade-off between performance and safety, reducing the unnecessary conservatism inherent in uncalibrated metrics. This necessitates a theoretical framework that directly links the learned safety critic to the probabilistic safety assessment of the system.

\subsection{Summary of Contributions}
In this work, we present \textbf{EigenSafe}, an operator-theoretic framework for accurate safety assessment of stochastic robotic systems. Within this framework, the evolution of the safety probability is viewed as a repeated application of a linear operator. The operator's dominant eigenpair, consisting of the dominant eigenvalue and eigenfunction, is shown to inherently capture the safety behavior of the closed-loop system over long time horizons.
Our contributions are summarized below.

\textbf{Spectral Safety Assessment.}
We develop a novel safety assessment method that can serve as an alternative to traditional optimal control-based approaches.
Its key advantage is that the derived spectral metrics are explicitly connected to the actual safety probability of the system, which is the most natural measure of safety. 
This theoretical grounding allows for a rigorous and unified assessment of safety, both locally for states and state-action pairs (via the dominant eigenfunction) and globally for the entire closed-loop system (via the dominant eigenvalue).

\textbf{Learning Algorithm.}
We propose a scalable, data-driven learning algorithm to approximate these spectral quantities from trajectory data without requiring an explicit model of the system dynamics.
Inspired by the power iteration algorithm, a classical method for finding the dominant eigenpair, we derive a loss functional that is minimized when the estimates converge to it. 
This allows the spectral safety critic to be learned using standard gradient descent.

\textbf{Application to Reinforcement Learning.}
We demonstrate the utility of EigenSafe in safe reinforcement learning (RL) tasks by formulating a safety-constrained policy optimization. This allows one to find a policy that balances well between maximizing the task rewards and maintaining long-term safety. Validation experiments are carried out in Gym \cite{2016openai} environments, showing comparable results both in terms of performance and safety to modern baselines. 

\textbf{Application to Imitation Learning.}
We apply EigenSafe to imitation learning (IL) by introducing a test-time safety filtering mechanism for learned stochastic policies. 
This ensures that only sufficiently safe input among multiple candidates is applied into the system, thereby achieving enhanced safety of the overall closed loop. We validate the approach in a real-world food preparation task using a UR3 robotic arm.

\section{Related Work}
We review relevant works on safety-critical control and their extensions to stochastic systems, as well as learning-based methods for safety critics and safe policies.

\textbf{Safety-Critical Control Theory. }
HJ reachability \cite{bansal2017hamilton, margellos2011hamilton} and CBFs \cite{ames2019control} are widely studied control methodologies for the safety of dynamical systems. 
In these frameworks, the reachability value function and the barrier function act as quantitative safety critics that measure the margin of safety for a given state with respect to a prescribed risk measure.
Closest to EigenSafe are stochastic extensions of CBFs. These extensions provide time-varying lower bounds on the safety probability, often in the form of exponentially decaying bounds \cite{clark2019control, cosner2023robust, jagtap2021formal, mathiesen2023safety, prandini2006stochastic, santoyo2021barrier, steinhardt2012finite, vinod2021stochastic, yaghoubi2020risk}. However, they are typically derived from conservative martingale inequalities, which can be overly conservative and often require specialized techniques to tighten the gap \cite{cosner2024bounding, jang2025upper}.
In practice, these safety-critical control methods are often deployed in the form of a \textit{safety filter} \cite{wabersich2023data}, whose role is to render a potentially unsafe reference policy safe by minimally modifying it \cite{ames2016control}.

\textbf{Safety Critic Learning. }
A growing body of work combines these control-theoretic methods with machine learning to synthesize safety critics \cite{brunke2022safe, dawson2023safe, ganai2024hamilton, gu2024review}.
Learning CBFs or HJ reachability value functions is relatively straightforward when the dynamics model is at least partially known \cite{bansal2021deepreach, choi2020reinforcement, liu2023safe, qin2022sablas, robey2020learning, so2024train}. However, this assumption can be unrealistic for complex or high-dimensional systems, such as systems subject to implicit safety constraints \cite{castaneda2023in, chou2018learning, kang2022lyapunov, kim2023learning, nakamura2025generalizing, seo2025uncertainty} or systems whose dynamics are represented through learned world models \cite{hafner2019learning, hafner2025mastering, micheli2023transformers, zhou2024dino}.
In these settings, model-free approaches based on RL \cite{fisac2019bridging, hsu2023isaacs, li2025certifiable} are better suited. Nevertheless, these approaches typically rely on optimal control formulations for deterministic systems and fail to capture the stochastic nature of real robotic systems. Moreover, because the learned critics are often tied to specific risk measures, they serve only as surrogates and are not calibrated to the actual safety probability. As a result, these safety critics tend to yield inconsistent safety assessments.
A few works propose HJ reachability formulations for safety probability, thereby correctly handling these issues \cite{ganai2023iterative, huh2020safe}. However, the infinite-horizon safety probability they consider is typically zero everywhere unless the system can be made almost-surely safe, which makes it difficult to compare relative safety across different states.

\textbf{Safe Decision Making.}
The goal of learning safety critics is to enable safe decision making. Safety filters are frequently employed with learned safety critics, often in the form of a switching filter, where the reference input is applied directly until the critic indicates a low safety level, after which the system falls back to a safe backup policy \cite{nakamura2025generalizing, seo2025uncertainty}.
However, this approach may lead to suboptimal task performance because it lacks a principled mechanism to balance safety and performance.
In safe RL, on the other hand, the decision-making problem is formulated as a constraint Markov decision process (CMDP), where the objective is to maximize rewards subject to safety constraints \cite{brunke2022safe, garcia2015comprehensive, gu2024review}. This safety constraints are commonly enforced by the use of Lagrange multipliers 
\cite{achiam2017constrained, ganai2023iterative} or epigraph reformulations \cite{so2023solving, zhang2025solving}.
Since the constraints in such optimizations are naturally defined with respect to the learned safety critic, their performance heavily depends on its quality. 
When the safety metric is poorly calibrated, these algorithms often produce overly conservative behavior.

\section{Problem Setup}

\subsection{Safety-Critical Dynamics}

The robotic system considered in this paper is modeled as a discrete-time Markov decision process (MDP) without the reward term, a 3-tuple $(S,A,P)$, where $S$ and $A$ are the state and action spaces, respectively, and $P$ is the transition probability that describes the dynamics:
\begin{equation} \label{eq: dynamics}
    s_{t+1} \sim P(\cdot | s_t, a_t).
\end{equation}
Here, $s_t, s_{t+1} \in S$ are the system's state at time $t$ and $t+1$, respectively, and $a_t \in A$ is the action taken at time $t$.

We consider a safety constraint $s_t \in C$, where $C \subsetneq S$ is the \textit{safe set}, i.e., the set of allowable states. For example, $C$ could represent the set of states satisfying a collision avoidance constraint.
Once the system leaves $C$ for the first time, a permanent safety failure is recorded, and the subsequent behavior is considered unsafe.
We model this using the notion of a \textit{termination state} $K\notin C$.
If $s_t \notin C$, then the trajectory is marked as stopped and terminated by setting $s_{t'} = K$ for all $t' \geq t$, regardless of the action taken thereafter. In this paper, we do not distinguish between different failure modes, so we can assume there exists only one unsafe state and write the state space as $S = C\cup\{K\}$.

The safety probability of the system is defined as the probability of the system not reaching the termination state $K$ throughout the given horizon, i.e., $\mathbb{P}[s_t \neq K]$.
This probability is naturally non-increasing in time (i.e., $\mathbb{P}[s_{t+1} \neq K] \leq \mathbb{P}[s_t \neq K]$) and depends on the chosen (feedback) policy and the initial state. 
The problem of interest in this paper is to derive a measure that quantitatively assesses the safety of a given state, action, or a feedback policy, in terms of how this safety probability evolves over a long time horizon.

\subsection{Dynamic Programming for Safety Probability}\label{subsec:dp_safety_probability}

Let $\pi$ be either a deterministic or a stochastic policy satisfying the Markov property. This means that the probability distribution of the action taken at time $t$ depends only on the current state $s_t$, i.e., $a_t \sim \pi(\cdot | s_t)$, for all $t \geq 0$.
Under this policy, we define $Z_\pi(t, x)$ as the probability that the system stays in the safe set until time $t$ given the initial state $s_0 = x\in S$:
\begin{equation}
    Z_\pi(t,x) \coloneqq \mathbb{P}_\pi [s_t \neq K | s_0 = x],
\end{equation}
where $\mathbb{P}_\pi$ denotes the probability measure induced by the closed-loop dynamics under the policy $\pi$.

It follows directly from the definition that $Z_\pi(0,x) = 1_C(x)$, where $1_C : S\rightarrow \{0,1\}$ is the indicator function of $C$.
From the law of total probability and the Markov property, one can easily see that the $(t+1)$-step safety probability from the initial condition $s_0 = x$, denoted $Z_\pi(t+1,x)$, can be expressed as the expectation of the $t$-step safety probability from the next state $s_{1}=x'$, $Z_\pi(t,s_{1})$. This yields the following dynamic programming principle: for any policy $\pi$, $t \in \{0,1,\cdots,\}$, and $x \in S$, 
\begin{equation} \label{eq: dynamic programming for Z}
    Z_\pi(t+1, x) 
    = \mathbb{E}_{u \sim \pi(\cdot|x),\; x' \sim P(\cdot|x,u)}\left[ Z_\pi(t, x') \right].
\end{equation}

In this paper, we view this dynamic programming principle from the perspective of an \textit{operator} acting on the space of scalar functions on $S$.
Formally, we start by letting $(\mathcal{F}, \norm{\cdot}_{\infty})$ be the Banach space of bounded and measurable scalar functions on $S$, where $\norm{\cdot}_\infty$ is the supremum norm defined as $\norm{\beta}_\infty = \sup_{x \in S}|\beta(x)|$. The operator $T_\pi: D_T\subset\mathcal{F}\rightarrow \mathcal{F}$, which describes this dynamic programming principle, is defined as 
\begin{equation} \label{eq: T_pi definition}
    T_\pi \beta(x) \coloneqq \mathbb{E}_{u \sim \pi(\cdot|x),\; x' \sim P(\cdot|x,u)}[\beta(x')],
\end{equation}
where the domain of $T_\pi$ is
\begin{equation} \label{eq: D_T}
    D_T \coloneqq \left\{\beta \in \mathcal{F} \;\middle|\; \beta(K) = 0\right\},
\end{equation}
which inherits the Banach space structure from $\mathcal{F}$.
This $T_\pi$ admits an integral\footnote{The integrals hereafter should be summations in case of discrete state or action spaces.} form
\begin{equation} \label{eq: T_pi integral}
    T_\pi \beta(x) 
    = \int_{S} \beta(y) p_\pi(y | x) dy
    = \int_{C} \beta(y) p_\pi(y | x) dy,
\end{equation}
where $p_\pi(\cdot | x)$ is the probability measure on $S$ induced by the policy $\pi$, with probability density function $p_\pi(y | x) \coloneqq \int_{A} \pi(u | x) P(y | x,u) du$, which can also be understood as the density of the next state $y$ given the current state $x$ under policy $\pi$. Note that the second equality holds since $\beta(x) = 0$ for any $x$ outside $C$.

One can also find that the image of $T_\pi$ is a subset of $D_T$, making $T_\pi$ an endomorphism (a mapping from $D_T$ onto itself). 
This follows from the fact that the system cannot recover from an unsafe state; hence, 
$\beta \in D_T$ implies $T_\pi \beta(x) = 0$ for all $x \notin C$.
Combined with the initial condition $Z_\pi(0,x) = 1_C(x)$, the dynamic programming principle \autoref{eq: dynamic programming for Z} can now be expressed as a $t$-time repeated application of the operator $T_\pi$ to the indicator function:
\begin{equation} \label{eq: repeated Tpi}
    Z_\pi (t, x) = T_\pi^t 1_C(x) = \underbrace{T_\pi\circ \cdots \circ T_\pi}_{\text{$t$ times}} 1_C(x).
\end{equation}

\section{Spectral Theory for Safety Assessment}
\subsection{Spectral Properties of $T_\pi$}

The operator $T_\pi$ can be understood as the restriction of the Koopman operator \cite{brunton2022modern, shi2026koopman} of the closed-loop system to the domain $D_T$ which is a vector subspace of the set of all measurable scalar functions on $S$.
It is evident from the definition \autoref{eq: T_pi definition} that regardless of $\pi$, $T_\pi$ is a \textit{linear} operator.
That is, if $\beta_1$ and $\beta_2$ are two scalar functions on $S$ and $c_1$ and $c_2$ are scalars, then $T_\pi(c_1\beta_1 + c_2 \beta_2) = c_1 T_\pi\beta_1 + c_2 T_\pi \beta_2$.
This allows one to examine its spectral properties.
Let the complex extension of the domain $D_T$ of $T_\pi$ be defined as $\overline{D}_T \coloneqq D_T + \mathrm{i} D_T = \{\beta_r + \mathrm{i}\cdot \beta_i\;|\; \beta_r, \beta_i \in D_T\}$.
We define an \textit{eigenpair} $(\gamma, \phi)$ as a pair of a scalar $\gamma \in \mathbb{C}$ and a nonzero function $\phi \in \overline{D}_T \setminus \{0\}$ satisfying
\begin{equation}
    T_\pi \phi(x) = \gamma \phi(x), \quad \forall x \in S.
\end{equation}
Since the domain $\overline{D}_T$ is typically infinite-dimensional (unless $S$ is a discrete set), there are typically infinitely many such pairs. We denote the eigenvalue with the biggest absolute value as the \textit{dominant eigenvalue}, $\gamma_\pi$, and its associated eigenfunction $\phi_\pi$ as the \textit{dominant eigenfunction}. The eigenpair $(\gamma_\pi, \phi_\pi)$ is referred to as the \textit{dominant eigenpair}.

One key characteristic of $T_\pi$ is that it is a \textit{positive operator}; that is, it maps nonnegative real functions to nonnegative real functions. 
This allows the use of Perron-Frobenius theory (and its infinite-dimensional generalization by Krein and Rutman \cite{krein_rutman}) to certify that $\gamma_\pi$ is a positive real number and that $\phi_\pi$ can be chosen to be nonnegative real everywhere on $S$.\footnote{This requires the assumption that $T_\pi$ is compact and the process is irreducible on the safe set $C$ (i.e., any state in $C$ is reachable from any other state in $C$ at a nonzero probability prior to arriving at $K$), which usually holds for realistic robotic systems if $C$ is a connected set.}

Moreover, $T_\pi$ is non-expansive with respect to the supremum norm $\norm{\cdot}_\infty$. For any $x \in C$, we observe
\begin{equation}
\begin{aligned}
    |T_\pi \beta(x)| &= \left|\int_C \beta(y) p_\pi (y|x) dy\right|
    \leq \norm{\beta}_\infty.
\end{aligned}
\end{equation}
This inequality implies that the spectral radius of $T_\pi$ is bounded by $1$; thus, all eigenvalues, including $\gamma_\pi$, should have absolute values less than or equal to $1$. Combining this with the positivity property mentioned above, we conclude that the dominant eigenvalue is a real number satisfying $0 \leq \gamma_\pi \leq 1$.

\subsection{The Dominant Eigenpair as a Safety Measure} \label{sec: safety value function}

Recall that the primary objective of this paper is to analyze the long-term safety of the closed-loop system, specifically, how the safety probability $Z_\pi$ of a given system evolves over long time horizons.
Once the dynamic programming recursion \autoref{eq: repeated Tpi} is expressed as repeated application of a linear operator, it is governed by the spectral structure of $T_\pi$,  with eigenmodes propagated multiplicatively according to their associated eigenvalues. In this context, the dominant eigenmode is of particular significance because the influence of non-dominant modes decays over time and eventually become negligible relative to the dominant eigenmode.

Usually, the dominant eigenvalue $\gamma_\pi$ has multiplicity $1$, meaning that the dominant eigenfunction $\phi_\pi$ is unique up to scalar multiplication. Assuming a nonzero spectral gap (separation between the modulus of the dominant eigenvalue and the rest of the spectrum), the influence of non-dominant modes decays geometrically faster than $\gamma_\pi^t$ as $t$ grows. Thus, for sufficiently large $t$, $Z_\pi(t,x)$ can be approximated as
\begin{equation} \label{eq: prob approximation}
    Z_\pi(t,x) = c \cdot \phi_\pi (x) \cdot \gamma_\pi^t + o(\gamma_\pi^t),
\end{equation}
where $c \in \mathbb{R}$ is a constant that does not depend on $t$. The additive term $o(\gamma_\pi^t)$ indicates that the approximation error decays strictly faster than $\gamma_\pi^t$. The approximation \autoref{eq: prob approximation} is therefore the tightest among exponential ones possible for the safety probability $Z_\pi$.

This approximation \autoref{eq: prob approximation} implies that the asymptotic decay rate of $Z_\pi$ is equal to $\gamma_\pi$ and is uniform across the entire state space.
Therefore, the dominant eigenvalue serves as a global metric for the safety of the closed-loop system under policy $\pi$, with a value closer to $1$ indicating a safer policy.
The value of the dominant eigenfunction $\phi_\pi(x)$ quantifies the relative safety of a specific state point $x \in C$ compared to other states in $C$. 
If this $\phi_\pi$ is chosen to be a nonnegative function, then a larger value of $\phi_\pi(x)$ corresponds to a higher degree of safety for that state $x$.

\begin{figure}
    \centering
    \includegraphics[width=\linewidth, page=1]{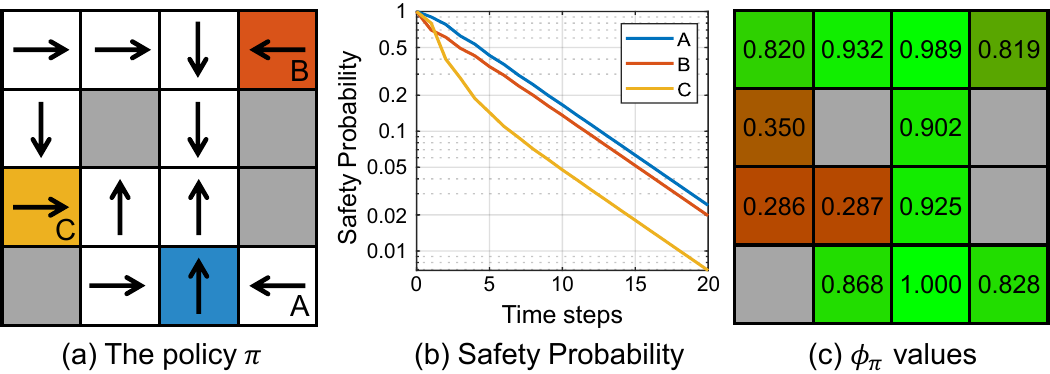}
    \vspace{-1.5em}
    \caption{A toy example describing the meaning of the dominant eigenpair of $T_\pi$. (a) This toy example consists of a finite state space represented by square cells, and a finite action space represented by an arrow on each of them.
    At each time step, the system moves to the adjacent cell in the arrow direction with probability $0.6$, and moves to each of the other adjacent cells or remains in the current cell with probability $0.1$. If the system enters a gray cell or leaves the map, it is deemed unsafe and is therefore considered to have reached the unsafe state $K$.
    (b) The safety probability given initial conditions A, B, C, corresponding to the colored cells in (a). Note that the vertical axis is log-scale and the slopes of all three curves converge to the same value, which corresponds to the dominant eigenvalue $\gamma_\pi$.
    (c) The dominant eigenfunction values. It can be seen that cells with higher values correspond to safer points. For finite-state systems, one can directly perform eigendecomposition of the finite-dimensional matrix representation of $T_\pi$ to obtain the eigenpair.}
    \vspace{-1em}
    \label{fig: toy for T}
\end{figure}

Collectively, these properties establish the dominant eigenpair $(\gamma_\pi, \phi_\pi)$ as a quantitative safety measure for the closed-loop system. These spectral quantities are intrinsically tied to the actual safety probability, which is the most natural indicator for safety.
The eigenvalue $\gamma_\pi$ is a safety measure for the closed-loop system itself or the policy, $\phi_\pi$ is the safety measure for the given state under $\pi$. See \autoref{fig: toy for T} (a) for a simple illustrative example using a simple finite-state system.

\subsection{The Safety Q Function}
The dominant eigenfunction $\phi_\pi$, which depends only on the state, provides a state-wise safety metric and, as such, plays a role analogous to a \textit{safety value function}. However, it cannot capture how safety varies with the action taken at a given state. In this subsection, we aim to derive a safety measure that provides a \textit{directional guide} for policy improvement. Achieving this requires a safety metric defined over state-action pairs, that is, a \textit{Q function} for safety, which extends the utility of $\phi_\pi$.

To this end, we broaden our discussion from the operator $T_\pi$ to assess safety at the level of state-action pairs, rather than solely at the state level. We introduce $H_\pi:S\times A \rightarrow \mathbb{R}$ as the safety probability conditioned on the initial state and action:
\begin{equation}
    H_\pi(t, x, u) \coloneqq \mathbb{P}_\pi \left[s_t \neq K \middle| s_0 = x, a_0= u \right],
\end{equation}
where $\mathbb{P}_\pi$ here measures the probability of the event assuming $a_k \sim \pi(\cdot | s_k)$ for all subsequent steps $k \in \{1,\cdots,t-1\}$. Similar to $Z_\pi$, $H_\pi$ takes nonnegative values upper-bounded by $1$ and satisfies the consistency condition
\begin{equation}
    \mathbb{E}_{u\sim \pi(\cdot|x)} [H_\pi(t,x,u)] = Z_\pi(t,x).
\end{equation}
Functionally, $H_\pi$ represents the safety probability of an \textit{augmented autonomous system} with state $\hat{s}_t \coloneqq (s_t, a_t)$, governed by the dynamics
\begin{equation} \label{eq: augmented system}
    s_{t+1} \sim P(\cdot | s_t, a_t), \quad
    a_{t+1} \sim \pi(\cdot | s_{t+1}),
\end{equation}
whose safety characteristics are the same as those of the original closed-loop system under policy $\pi$, except for the initial step $t=0$.

\begin{figure}
    \centering
    \includegraphics[width=\linewidth]{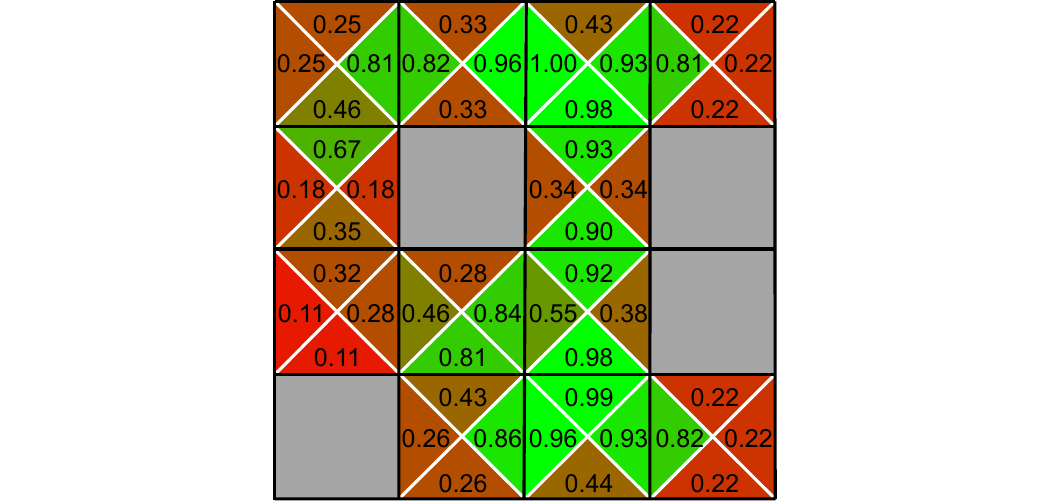}
    \vspace{-1.5em}
    \caption{The dominant eigenfunction $\psi_\pi$ of $A_\pi$, defined with respect to the same dynamics and $\pi$ as in \autoref{fig: toy for T}. Similar to \autoref{fig: toy for T}~(c), the eigenfunction can be computed by directly eigendecomposing the matrix representation of $A_\pi$.
    Each square represents a state, and the numbers inside the triangles denote the values of $\psi_\pi$ for the specific state-action pair defined by the square and the triangle's direction.
    It can be seen that the eigenfunction assigns higher values to safer transitions.}
    \vspace{-1.5em}
    \label{fig: toy for A}
\end{figure}

Similar to $T_\pi$, we define the operator
\begin{equation}
    A_\pi \beta(x, u) \coloneqq \mathbb{E}_{x' \sim P(\cdot |x,u), u' \sim \pi(\cdot | x')} \left[\beta(x', u')\right]
\end{equation}
on the domain $D_A \coloneqq \left\{\beta\in \mathcal{G} \;\middle|\; \beta(K, u) = 0, \; \forall u \in A \right\}$,
where $(\mathcal{G},\norm{\cdot}_\infty)$ is the Banach space of bounded and measurable scalar functions on $S\times A$, equipped with the supremum norm.
This allows us to follow the same procedure in \autoref{sec: safety value function} to derive the dynamic programming principle for $H_\pi$:
\begin{equation}
    H_\pi(t+1, \cdot, \cdot) = A_\pi H_\pi(t,\cdot, \cdot),
\end{equation}
which implies $H_\pi(t,x,u) = A_\pi^t 1_{C\times A}(x, u)$.

It is evident that $A_\pi$ is also a nonnegative and non-expansive linear operator. Since the augmented system shares the underlying dynamics of the original system, the eigenstructure of $A_\pi$ is closely related to that of $T_\pi$.
Specifically, there exists a dominant eigenpair $(\gamma_\pi, \psi_\pi)$ such that $A_\pi \psi_\pi(x,u) = \gamma_\pi \psi_\pi(x,u)$ for all $(x, u) \in S \times A$. The dominant eigenvalue $\gamma_\pi$ is identical to that of $T_\pi$ and the dominant eigenfunction $\psi_\pi$ is a nonnegative function on $S \times A$. Furthermore, $\psi_\pi$ relates to $\phi_\pi$ through the expectation over the closed-loop actions 
\begin{equation}
    \mathbb{E}_{u \sim \pi(\cdot|x)} \psi_\pi(x,u) = c\cdot \phi_\pi(x), \quad \forall x \in C,
\end{equation}
for an appropriate constant $c \in \mathbb{R}$ not depending on the state.

Similarly to \autoref{eq: prob approximation}, one can approximate $H_\pi$ as
\begin{equation} \label{eq: H approximation}
    H_\pi (t,x,u) = A_\pi^t 1_{C\times A} (x,u) = c \cdot \psi_\pi(x,u) \cdot \gamma_\pi^t + o(\gamma_\pi^t),
\end{equation}
where $c \in \mathbb{R}$ is an appropriate constant.
This suggests that the eigenfunction $\psi_\pi$ quantifies the safety level of a given state-action pair $(x,u)$. Just as $\phi_\pi$ plays a role as a safety value function, $\psi_\pi$ serves as a safety Q function, providing a metric to guide policy improvements during training. We illustrate the role of the eigenfunction $\psi_\pi$ in \autoref{fig: toy for A} using the same tabular dynamics as in \autoref{fig: toy for T}.

\section{EigenSafe: Learning the Eigenpair} \label{sec: eigenpair learning}

In this section, we present a practical methodology for learning the safety Q function $\psi_\pi$ and its associated eigenvalue from transition data, building upon the theoretical framework established in the previous section.

\subsection{Eigenpair Learning}

Consider a dataset, batch, or a replay buffer $\mathcal{D}$ comprising transition tuples $(x,u,x')$ of state, action, and the next state. In its simplest form, EigenSafe parametrizes the eigenvalue $\gamma$ as a learnable real scalar and approximates the eigenfunction $\psi$ using a function approximator, such as a lookup table for discrete spaces or a neural network for continuous spaces.
The parameters are optimized by minimizing the following loss functional:
\begin{equation} \label{eq: eig loss}
\begin{multlined}
    \mathcal{J}_\text{eig}[\gamma, \psi] \coloneqq \\ \frac{W_\text{eig}}{|\mathcal{D}|}\sum_{(x,u,x') \in \mathcal{D}} \left(
    \begin{aligned}
        1[x \in C \wedge x' \in C]\cdot \psi(x', u') \quad \quad \\ 
        {} - \gamma \cdot 1[x \in C] \cdot \psi(x,u)
    \end{aligned}\right)^2 \\
    {} + W_n \cdot \left( \max_{(x,u,\cdot) \in \mathcal{D}} \psi(x,u) - 1 \right)^2,
\end{multlined}
\end{equation}
where $u'$ in the summation is sampled from the policy conditioned on the next state $x'$ from the dataset, and $W_\text{eig},\; W_n >0$ are tunable hyperparameters.
The indicator function $1[\cdot]$ evaluates to $1$ if the condition inside the bracket holds, and $0$ otherwise.
The first term in \autoref{eq: eig loss} constrains $\psi$ to be zero outside the safe set $C$, while ensuring it satisfies the eigenvalue equation $A_\pi \psi(x,u) = \gamma\psi(x,u)$ within $C$.
The second term serves as a soft normalization constraint that ensures the learned eigenfunction maintains a unit supremum norm, $\|\psi\|_\infty = 1$, and prevents converging to a degenerate solution $\psi=0$.

While standard Q learning algorithms from RL utilize a discount factor strictly smaller than $1$ to ensure a unique solution via contraction mapping theory, the loss functional \autoref{eq: eig loss} does not theoretically admit a unique minimizer. This is because \textit{any} valid eigenpair of the operator $A_\pi$, including the non-dominant ones, satisfies the eigenvalue equation enforced by \autoref{eq: eig loss}. Instead, the convergence of the proposed approach is grounded in the principles of the \textit{power iteration} algorithm, also known as the von Mises iteration.
To determine the dominant eigenpair of a positive linear operator $L:V\rightarrow V$, power iteration initializes with a nonzero initial guess $v \in V \setminus \{0\}$ and recursively applies the update
\begin{equation} \label{eq: power iteration}
    \lambda \leftarrow \norm{Lv}, \quad
    v \leftarrow Lv / \norm{Lv}.
\end{equation}
Although this update rule is not a contraction mapping in general, the pair $(\lambda, v)$ converges to the dominant eigenpair of $L$ with probability $1$, provided that the dominant eigenvalue has multiplicity $1$ and is strictly bigger than the modulus of every other eigenvalue.
The normalization step ensures that the magnitudes of non-dominant modes eventually decay relative to the dominant mode, isolating the dominant eigenpair.
While RL-based approaches to learning safety critics rely on artificial discount factors (which can undermine the theoretical validity of the guarantees they provide) to ensure convergence, EigenSafe achieves robust empirical convergence without the need for such modifications.

\subsection{Remarks on Implementation} \label{sec: eigensafe remarks}

Empirically, minimizing $\mathcal{J}_\text{eig}$ alone is usually sufficient for convergence, regardless of initialization. However, to accelerate training during the initial phase, one may optionally choose to employ the following additional loss term
\begin{equation} \label{eq: positivity loss}
    \mathcal{J}_+[\psi] = \frac{W_+}{|\mathcal{D}|}\sum_{(x,u,\cdot) \in \mathcal{D}} \operatorname{ReLU}\left(-\psi(x,u)\right)
\end{equation}
in addition to $\mathcal{J}_\text{eig}$, where $W_+$ is a tunable nonnegative weight.
This term penalizes negative values and thereby biases the optimization toward nonnegative $\psi$, reducing the influence of sign-oscillatory components associated with non-dominant modes.
Once the dominant eigenmode becomes prominent, this auxiliary loss quickly vanishes, not affecting the overall learned result.

In cases where the policy $\pi$ is only implicitly known through the trajectory dataset itself, 
and thus the next action $u'$ cannot be actively sampled from $\pi(\cdot|x')$, 
one can substitute it with the action $u'$ recorded in the dataset trajectory $(x,u,x',u')$. In this case, $\gamma$ and $\psi$ will converge to the dominant eigenpair corresponding to the \textit{behavior policy} that generated the dataset.

Finally, consistent with standard practices in deep RL, we employ a semi-gradient approach to enhance training stability. Specifically, during training, we treat the term $\psi(x',u')$ in \autoref{eq: eig loss} as a fixed regression target during the optimization step.
This is achieved by detaching the calculated values of $\psi(x',u')$ from the computational graph where gradient is propagated, effectively mitigating the moving target problem. 

\begin{figure}
    \centering 
    \includegraphics[width=\linewidth]{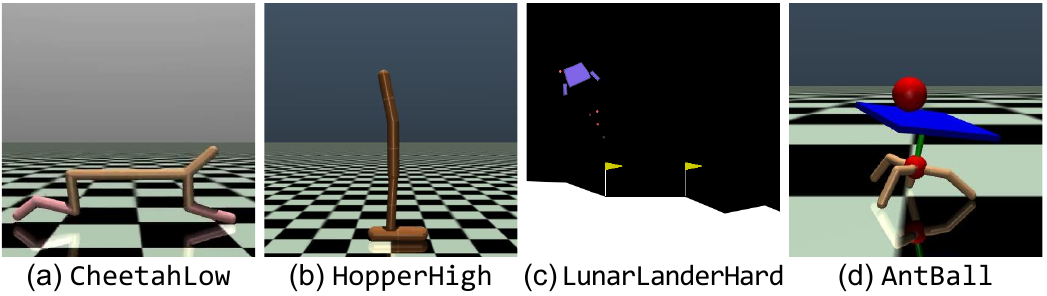}
    \vspace{-1.5em}
    \caption{The Gym environments used in the safe RL demonstration.}
    \vspace{-0.5em}
    \label{fig: safe rl environments}
\end{figure}

\section{Safe Reinforcement Learning using EigenSafe} \label{sec: rl}

In this section, we demonstrate the application of the learned eigenfunction as a safety Q function in an online safe RL framework. We present experimental results obtained from Gym environments \cite{2016openai}.

\subsection{Setup}

The objective of this safe RL setting is to find a policy $\pi$ that maximizes the expected return (the discounted sum of rewards) while satisfying the safety constraint, ensuring that the agent remains within the safe set $C$ with a probability above a specified threshold.
To achieve this, we impose a spectral constraint on the policy: 
\begin{equation} \label{eq: eigenvalue constraint}
    \gamma_\pi \geq \gamma_0,
\end{equation}
where $\gamma_0 \in [0,1]$ is the target eigenvalue typically chosen to be very close to (and practically equal to) $1$. As $\gamma_\pi$ is a global indicator for the safety of $\pi$, this informally interprets to: the policy $\pi$ should be \textit{safer} than the given threshold.

We adopt the soft actor-critic (SAC) \cite{haarnoja2018soft} framework which utilizes a stochastic policy that facilitates exploration. The proposed EigenSafe framework is naturally compatible with the stochastic transitions inherent in SAC. The task critic $Q_\pi$ is trained via the standard Bellman loss to match the entropy-regularized expected return. Simultaneously, the eigenpair $(\gamma_\pi, \psi_\pi)$ is learned by minimizing the loss functional introduced in \autoref{eq: eig loss}.

The challenge in training the policy $\pi$ is that the eigenvalue $\gamma_\pi$ is a global property of the operator $A_\pi$.
It depends on the policy only implicitly and does not directly relate to the local state-action inputs, making gradient-based optimization intractable. To address this, we replace the global constraint \autoref{eq: eigenvalue constraint} with an equivalent local condition
\begin{equation} \label{eq: eigenvalue constraint transformed}
    \mathbb{E}_{x' \sim P(\cdot | x,u), u' \sim \pi(\cdot|x')} \psi_\pi (x',u') \geq \gamma_0 \psi_\pi(x,u),
\end{equation}
which should be satisfied for all $(x,u) \in S\times A$.
If a policy $\pi$ satisfies \autoref{eq: eigenvalue constraint transformed} for all state-action pairs, then the global eigenvalue condition is satisfied, and vice versa.\footnote{This equivalence is the infinite-dimensional analogue of the Collatz-Wielandt formula for positive matrices \cite{meyer2023matrix}.} The policy optimization can therefore be formulated as the following constrained problem:
\begin{equation} \label{eq: rl loss original}
\begin{aligned}
    \max_{\pi} \quad &
    \mathcal{J}_\text{RL} [\pi] \coloneqq \mathbb{E}_{(x,\cdot, \cdot) \sim \mathcal{D}, u \sim \pi(\cdot|x)} \left[Q_\pi (x, u)\right] \\
    \operatorname{s.t.}\quad & \mathbb{E}_{x' \sim P(\cdot | x,u), u' \sim \pi(\cdot|x')} \psi_\pi (x',u') \geq \gamma_0 \psi_\pi(x,u),
\end{aligned}
\end{equation}
where the constraint should be satisfied for all $(x, u, \cdot)$ drawn from the replay buffer $\mathcal{D}$.

We solve this constrained optimization by reformulating its objective into an unconstrained augmented one using the method of Lagrange multipliers \cite{boyd2004convex, eysenbach2021robust}. The augmented objective is defined as
\begin{multline} \label{eq: rl loss augmented}
    \hat{\mathcal{J}}_\text{RL}[\Lambda, \pi] \coloneqq 
    \mathcal{J}_\text{RL}[\pi] + \mathbb{E}_{(x,u,x') \sim \mathcal{D}, u' \sim \pi(\cdot|x')} \big[ \\
    \Lambda(x,u)\cdot \min\left\{\epsilon, \psi_\pi(x',u') -  \gamma_0 \psi_\pi(x,u)\right\} \big].
\end{multline}
Here, $\Lambda(\cdot, \cdot)$ represents the state- and action-dependent Lagrange multiplier, and $\epsilon>0$ is a positive but very small constant employed to prevent $\Lambda$ from exploding towards positive infinity. The optimization is performed via the dual gradient descent algorithm. It simultaneously updates $\pi$ and $\Lambda$ to maximize and minimize, respectively, the augmented objective $\hat{\mathcal{J}}_{\text{RL}}$. This process yields a policy that maximizes the task reward while adhering to the safety constraints.

\subsection{Results}

\begin{figure}
    \centering
    \includegraphics[width=\linewidth]{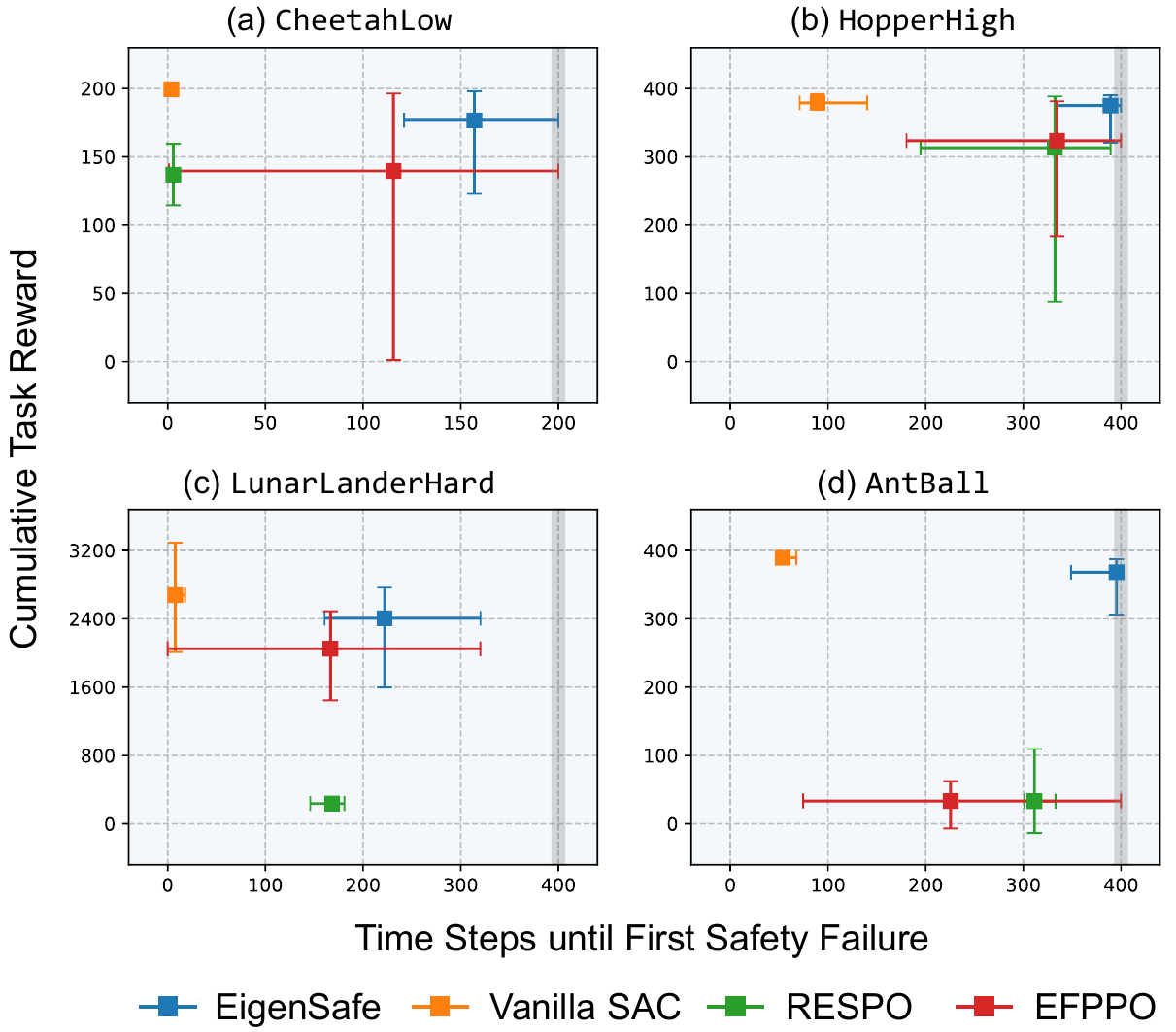}
    \vspace{-1.5em}
    \caption{Baseline comparison results for safe RL.
    The horizontal axis denotes the number of steps taken until a safety failure or the the agent has reached the maximum episode limit, while the vertical axis represents the total reward accumulated throughout the episode, regardless of the safety outcome.
    EigenSafe consistently appears in the upper-right region across all environments tested, indicating a better balance between reward and safety. Squares indicate mean values, and error bars denote maximum and minimum values over the rollouts. Gray vertical bars indicate the maximum episode length. Each performance is evaluated over five rollout episodes during the final three evaluation epochs, across four training seeds.}
    \vspace{-1.5em}
    \label{fig: rl performance}
\end{figure}

We demonstrate this online safe RL framework in four Gym environments as shown in \autoref{fig: safe rl environments}.
\begin{enumerate}[(a)]
    \item \texttt{CheetahLow} is based on the standard \texttt{HalfCheetah} environment. The agent receives reward based on its forward velocity only, and the constraint is to maintain its torso below a certain level.
    \item \texttt{HopperHigh} is based on the standard \texttt{Hopper} environment. The agent receives reward based on its forward velocity only, and the constraint is to maintain its torso above a certain level.
    \item \texttt{LunarLanderHard} is a more difficult variant of the standard \texttt{LunarLander} environment, with diversified initial reset states. 
    The reward is given only for reaching the goal, and the safety constraints are imposed on landing velocity, body orientation, and horizontal position.
    \item \texttt{AntBall} is a modified version of the \texttt{Ant} environment.
    The agent receives reward based on its forward velocity only, and the safety constraint is to not drop the ball from the plate attached to the torso.
\end{enumerate}
We compare with three baselines.
Vanilla SAC \cite{haarnoja2018soft} is included as an unconstrained RL baseline that purely maximizes the reward without accounting for safety constraints.
RESPO \cite{ganai2023iterative} learns a reachability value function that predicts future constraint violations.
The policy is trained through dual gradient descent using Lagrange multipliers.
EFPPO \cite{so2023solving} instead reformulates the safety-constrained RL problem into an epigraph form. It searches over a cost-budget variable to satisfy constraints, without using Lagrange multipliers.
As shown in \autoref{fig: rl performance}, the proposed method achieves a better safety-performance balance than the baselines, which we attribute to EigenSafe’s improved accuracy of safety assessment.

\begin{figure}
    \centering
    \includegraphics[width=\linewidth]{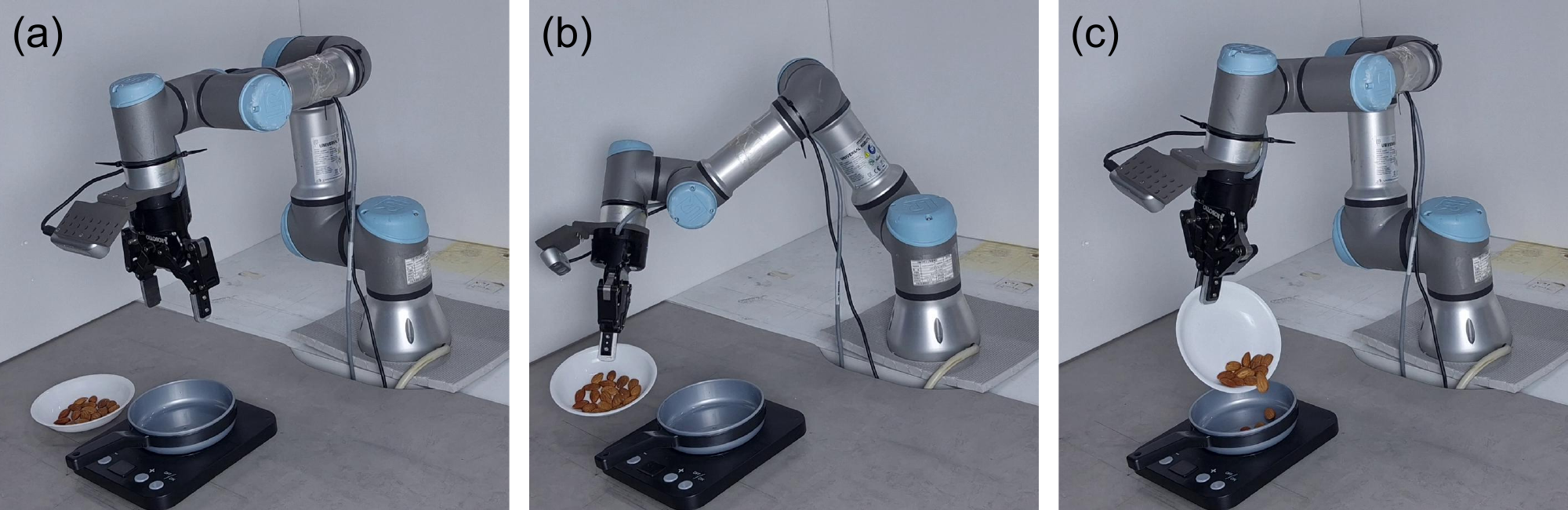}
    \vspace{-1em}
    \caption{
    The task for the hardware experiment in \autoref{sec: il} is to safely pick up the bowl of almonds and pour the almonds into the pan, without spilling them or causing a collision.
    }
    \label{fig: task description}
\end{figure}

\section{Safety-Filtered Inference for Imitation Learning using EigenSafe} \label{sec: il}

We demonstrate another practical application of the learned eigenfunction $\psi_\pi$, utilizing it as a safety Q function to perform test-time safety filtering on a learned stochastic policy.

\subsection{Setup}

We consider a standard behavior cloning (BC) setup.
The objective is to filter the output of a stochastic BC policy, selecting only those actions deemed relatively safe among multiple candidates.
This approach consists of two key components: (a) a learned safety metric (the eigenpair) that evaluates the safety of the chosen action,
and (b) an expressive stochastic policy that generates diverse action candidates reflecting the multi-modality of the demonstration data.
Unlike the online RL setting where the policy itself is updated, this task focuses on test-time safety improvement.
Instead of updating the policy weights, we bias the action sampling process of a pre-trained stochastic policy $\pi_\text{BC}$ to favor safer transitions.

To learn the eigenfunction we utilize trajectory tuples $(x,u,x',u')$ consisting of state, action, next state, and next action from the dataset.
The EigenSafe loss function \autoref{eq: eig loss} is minimized using the next action $u'$ recorded in the dataset, resulting in the learned eigenpair reflecting the safety characteristics of the behavior policy that generated the data.
At the same time, the stochastic policy $\pi_\text{BC}(\cdot|x)$ is trained using flow matching \cite{lipman2022flow, liu2022flow, park2025flow}, a generative modeling technique that ensures sampled actions cover the multi-modality of the training data.

\begin{algorithm}[t]
\caption{Safety-Filtered Inference for IL}\label{alg: il}
\begin{algorithmic}[1]
    \STATE \textbf{Input:} Pretrained stochastic policy $\pi_\text{BC}$, Pretrained eigenfunction $\psi_\pi$, Hyperparameters $n, k$
    \FOR{every time step $t$}
        \STATE Observe the current state $s_t$.
        \STATE Sample $n$ candidate actions $\{u^{(i)}\}_{i=1}^n \sim \pi_\text{BC}(\cdot|s_t)$.
        \STATE Evaluate safety scores $v^{(i)} \leftarrow \psi_\pi(s_t, u^{(i)})$ for all $i$.
        \STATE Select action $a_t = u^{(j)}$ such that $v^{(j)}$ is the $k$-th largest value among the safety scores.
        \STATE Execute action $a_t$.
    \ENDFOR
\end{algorithmic}
\end{algorithm}

During inference, we employ a safety-filtering mechanism to select the most appropriate action from the stochastic policy. At each time step $t$, given the current state $s_t$, we sample $n$ candidate actions $\{u^{(i)}\}_{i=1}^n$ from the trained BC policy $\pi_\text{BC}(\cdot|s_t)$.
We then evaluate the safety score of each candidate using the learned eigenfunction $v^{(i)} = \psi_\pi (s_t, u^{(i)})$.
The action corresponding to the $k$-th largest value among $v^{(i)}$-s is chosen as the input $a_t$ and applied to the system.
Here, $n$ and $k$ are tunable parameters.
For best performance, we avoid using the highest-value action ($k=1$) because it increases the risk of out-of-distribution (OOD) actions relative to the behavior policy. Empirically, setting $k \approx n/5$ yields the best performance in both task completion and safety.
The overall procedure is summarized in \autoref{alg: il}.

\begin{figure}
    \centering
    \includegraphics[page=1, width=\linewidth]{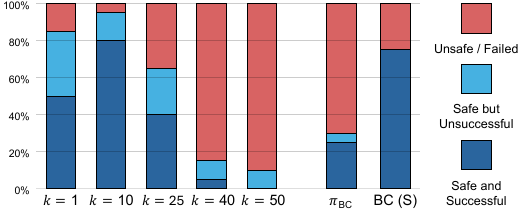}
    \caption{Success/safety rates in safety-filtered IL, with $n=50$. Except for the $k=1$ case, there is a positive correlation between selecting actions with higher $\psi_\pi$ values and the resulting success and safety rate. 
    Notably, the proposed safety-filtered approach achieves higher safety performance than even the flow BC policy trained only on successful demonstrations. 
    Dark blue bars represent full task success (pouring almonds safely), while light blue indicates episodes that remained safe (no violations) but failed to complete the task. BC (S) refers to the policy trained solely on successful demonstrations.
    The success and safety rates are estimated by conducting 20 repeated trials for each scenario.}
    \label{fig: il success rate}
\end{figure}

\begin{figure*}
    \centering
    \includegraphics[width=\linewidth]{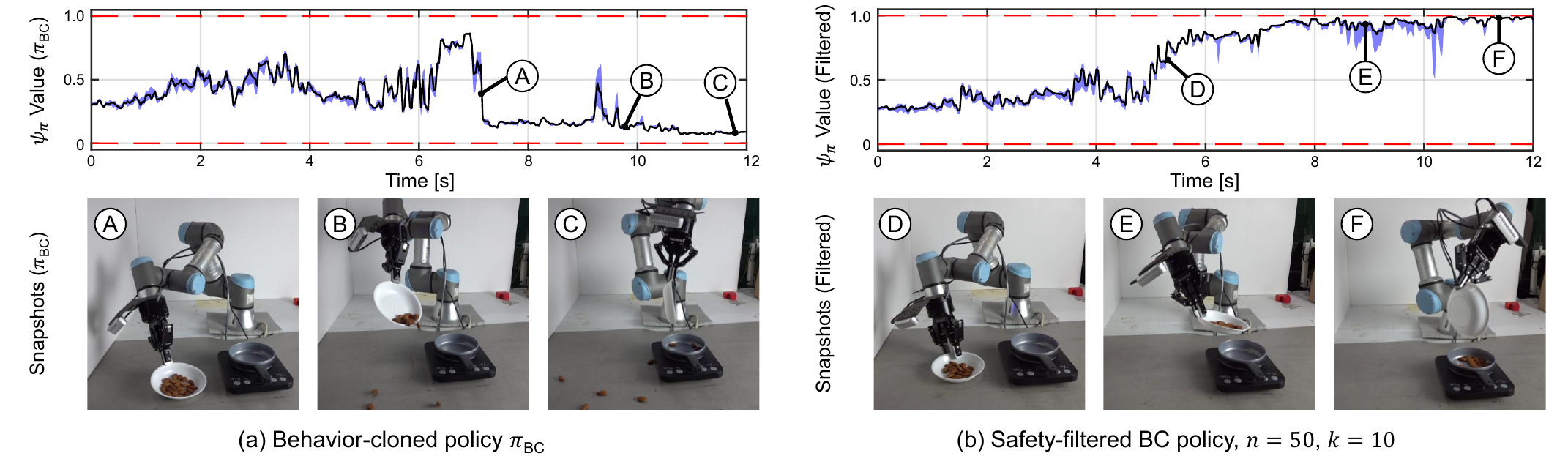}
    \vspace{-1.5em}
    \caption{Temporal evolution of the eigenfunction $\psi_\pi$ value during task execution. The blue shaded regions represent the range of eigenfunction values across the sampled action candidates at each time step. The black curves represent that of the chosen action. (a) The naive BC policy $\pi_\text{BC}$ leads to a safety violation. It can be seen that the eigenfunction value drops (at \textcircled{A}) ahead of the actual spillage (see \textcircled{B}), demonstrating the capability of the eigenfunction to predict future failures. (b) The safety-filtered policy (using $n=50$ and $k=10$) consistently selects actions with high safety scores. This ensures a more stable grasp \textcircled{D} compared to \textcircled{A}, allowing the robot to convey (\textcircled{E}) and pour (\textcircled{F}) the almonds safely. }
    \vspace{-1em}
    \label{fig: il result}
\end{figure*}

\subsection{Results} \label{sec: rl results}

We validate the method on a food preparation task depicted in \autoref{fig: task description}. The objective is to control the UR3 robot equipped with a Robitq 2F-85 gripper to grasp a bowl of almonds and pour them into a pan on a toy stove.
Any spill of almonds outside the pan or collision between the robot and the environment is considered a safety violation.
The robot's perception consists of two RGB cameras: one attached to the wrist of the arm, and one on an external mount that provides the third-person view of the workspace. The images are processed by a DINOv2 encoder \cite{oquab2023dinov2} to extract latent features, which are then concatenated with the robot's joint angles and gripper status, forming a 776-dimensional state.
The dataset was collected using a GELLO teleoperation device \cite{wu2024gello}. The dataset comprises 150 successful demonstrations, where the task was completed without safety violations, and 203 failed trajectories, which terminated due to spillage or collision.

In the experiments, we set the number of action candidates $n=50$ and evaluated performance across different selection of $k$, by recording the success and safety rates over 20 trials each. We also compare with the naive BC baseline $\pi_\text{BC}$ which is trained on the whole dataset, and another BC policy trained only on successful demonstrations. These rates are summarized in \autoref{fig: il success rate}.
It can be seen that there is a clear positive correlation between the eigenfunction value and both the success and safety rates, with the only exception being the $k=1$ case, which is affected by the OOD issue discussed above.
Notably, the optimal $k=10$ case exhibits better performance even compared to the BC trained only on successful demonstration, which is not aware of the safety boundary.
\autoref{fig: il result} visualizes the evolution of $\psi_\pi$ values for (a) our safety-filtered approach using an optimal rank $k=10$, and (b) the naive BC baseline.
It can be seen from the snapshots that the learned $\psi_\pi$ serves as an effective safety metric, assigning higher values to safer states.
Moreover, it also exhibits \textit{prediction} capabilities.
For instance, at point \textcircled{A}, the eigenfunction value drops significantly upon detecting a precarious grasp, accurately anticipating the eventual failure before it physically occurs.

\section{Conclusion}

\subsection{Summary}

This paper presents EigenSafe, a novel theoretical framework for learning safety value functions based on linear operator theory rather than optimal control. We derive a linear Bellman-like operator for dynamic programming for safety probability, whose dominant eigenpair provides a safety measure that is directly tied to the safety probability.
In particular, the dominant eigenvalue serves as a scalar metric quantifying the global decay rate of the system's safety probability, while the corresponding dominant eigenfunction serves as a safety Q function that quantifies the relative safety of state-action pairs and provides a directional guidance for safety improvement.
This dominant eigenpair can be learned via the proposed loss function inspired by the power iteration algorithm. 

We demonstrate the versatility of EigenSafe in safe RL and IL tasks.
For safe RL, we formulate a policy optimization problem that maximizes the expected return (the discounted sum of task rewards) subject to a lower-bound constraint on the dominant eigenvalue. Empirical results across various Gym environments confirm that the approach is effective in achieving both safety compliance and reward maximization.
For IL, we propose a test-time safety-filtering technique for a learned stochastic BC policy that evaluates the safety of multiple candidate actions using the learned eigenfunction and executes only those deemed safe. Experiments on a food preparation task using a UR3 robotic arm demonstrate that higher safety and success rates can be achieved with this filtering technique, compared to both naive BC policies and policies trained exclusively on successful demonstrations.

\subsection{Limitations and Future Work}

We conclude this paper by summarizing the limitations of the proposed approach and future research directions.

\textbf{Overestimation Bias in Online Learning.} 
In the online RL setting, where the policy and the eigenfunction are learned simultaneously, EigenSafe may overestimate the safety of certain areas before the policy learning has stabilized. While this is well-known and common in RL \cite{hasselt2010double}, it becomes particularly problematic in terms of safety. In the future work, we plan to apply techniques such as \cite{hasselt2010double, kumar2020conservative} to mitigate this issue.

\textbf{Dependency on Failure Data and Out-of-Distribution Risks.} 
The current framework fundamentally relies on failure labels, i.e., transitions leaving the safe set $C$. Consequently, the method might struggle to distinguish between the true safe regions and OOD regions where data is scarce, especially in the offline learning setting.
Future work will address this by integrating OOD avoidance into the safety constraints,
e.g., \cite{castaneda2023in, kang2022lyapunov, seo2025uncertainty}, or by introducing additional penalties near the boundary of the training dataset.

\textbf{Future Research Directions.}
In addition to addressing these limitations, future research will focus on exploiting prior information to enhance sample efficiency and reduce the reliance on failure experiences. We believe one can leverage large foundation models to infer common-sense safety constraints and physical priors, enabling the agent to recognize unsafe areas with less failure experiences.
In addition, acknowledging that the definition of safety might vary across domains and may not be fully captured by probability alone, we plan to generalize the framework to more general risk metrics, for example, conditional value at risk, allowing the framework to address different severity of failure modes.

% \newpage

\section*{Acknowledgments}
The authors would like to thank Prof. Jason J. Choi (UCLA) and Prof. Namhoon Cho (Seoul National University) for insightful discussions.
Jonghun Shin, Jusuk Lee, and Seungwon Roh (Seoul National University) contributed in the hardware experiment setup.
This work was supported by Samsung Research Funding \& Incubation Center of Samsung Electronics under Project Number SRFC-IT2402-17. The work of Inkyu Jang was partially supported by the Basic Science Research Program through the National Research Foundation of Korea (NRF) funded by the Ministry of Education (RS-2024-00407121).

%% Use plainnat to work nicely with natbib. 

\bibliographystyle{plainnat}
\bibliography{references}

\newpage

\appendices

\counterwithin{equation}{section}
\counterwithin{figure}{section}
\counterwithin{table}{section}
\counterwithin{theorem}{section}
\counterwithin{lemma}{section}
\counterwithin{proposition}{section}
\counterwithin{assumption}{section}
\counterwithin{corollary}{section}

\section{Formal Analysis and Mathematical Guarantees}

This section provides mathematical proofs for the properties of the operators $T_\pi$ and $A_\pi$ and their dominant eigenpairs $(\gamma_\pi, \phi_\pi)$ and $(\gamma_\pi, \psi_\pi)$. While the proofs below are mainly described for $T_\pi$ only, the same techniques can also be applied to $A_\pi$.
We introduce appropriate assumptions on the closed-loop dynamics in order for the guarantees to hold, and discuss their practicality.
\autoref{tab:theory_guide} provides a high-level view on the implications of the theorems in this appendix for EigenSafe.

\subsection{Background on Operator Theory}
We start with a brief introduction to linear operator theory and functional analysis, which are the key mathematical concepts used throughout the paper. Further details can be found in established textbooks such as \cite{conway2019course, naylor1982linear}.

Consider a linear operator $L:\mathcal{X}\rightarrow \mathcal{X}$, where $\mathcal{X}$ is a complex Banach space equipped with the norm $\norm{\cdot}$.
Let $\operatorname{id}:\mathcal{X}\rightarrow \mathcal{X}$ be the identity map, and $L_z \coloneqq L - z\cdot \operatorname{id}$ for a $z \in \mathbb{C}$.
The value $z$ is called a \textit{regular value} of $L$ if the following conditions are met:
\begin{itemize}
    \item $L_z$ is an injective map.
    \item The range of $L_z$, $L_z(\mathcal{X})$, is a dense subset of $\mathcal{X}$. 
    \item The inverse $L_z^{-1}$, defined on $L_z(\mathcal{X}) \subseteq \mathcal{X}$, is a bounded operator, meaning there exists a $M>0$ such that $\norm{L_z^{-1}x} \leq M\norm{x}$ for all $x$.
\end{itemize}
The set of all regular values of $L$, $R(L)$ is called the \textit{resolvent set}. The complement of the resolvent set, $\mathbb{C} \setminus R(L)$, is called the \textit{spectrum} of $L$, written as $\sigma(L)$.

If a nonzero vector $x \in \mathcal{X}\setminus \{0\}$ satisfies
\begin{equation}
    Lx = \lambda x
\end{equation}
for a scalar $\lambda \in \mathbb{C}$, $x$ is called an \textit{eigenvector} of $L$, and $\lambda$ is called the \textit{eigenvalue}. In particular, $x$ is also called an \textit{eigenfunction} if $\mathcal{X}$ is the vector space of functions.
Since $L_\lambda = L - \lambda \cdot \operatorname{id}$ is not invertible, every eigenvalue belongs to the spectrum $\sigma(L)$. The converse is also true for finite-dimensional $\mathcal{X}$, but in infinite dimensions, $\sigma(L)$ might contain elements that are not necessarily eigenvalues.

The spectral radius of $L$, $\rho(L)$, is defined as the supremum of the absolute values of the elements in $\sigma(L)$, i.e., $\rho(L) = \sup_{s \in \sigma(L)} |s|$.
It is a known fact that if $L$ is a bounded operator such that $\norm{Lx} \leq M\norm{x}$ for all $x \in \mathcal{X}$, then $\rho(L) \leq M$. This can be obtained using the Gelfand's formula
\begin{equation}
    \rho(L) = \lim_{n\rightarrow \infty} \norm{L^n}^{\frac{1}{n}},
\end{equation}
where $\norm{\cdot}$ here is the operator norm defined as
\begin{equation} \label{eq: operator norm}
    \norm{L} = \sup_{x \in \mathcal{X} \setminus \{0\}} \frac{\norm{Lx}}{\norm{x}}.
\end{equation}

A linear operator $L:\mathcal{X}\rightarrow \mathcal{Y}$ between two Banach spaces is said to be compact if it maps bounded subsets of $\mathcal{X}$ to relatively compact subsets of $\mathcal{Y}$. Here, \textit{relatively compact} means that the closure of the image is compact.
If $L$ is a mapping between finite-dimensional vector spaces, $L$ is always compact. This easily follows from the fact that in Euclidean space, compactness is equivalent to being bounded and closed. However, in infinite dimensions, this is not always true.

Compactness ensures that the spectral properties of an infinite-dimensional operator behaves similarly to a finite-dimensional matrix. 
\begin{theorem}[Spectral Theorem for Compact Operators] \label{thm: spectral theorem}
    If $L:\mathcal{X}\rightarrow \mathcal{X}$ is a compact linear operator on a Banach space $\mathcal{X}$, then the spectrum $\sigma(L)$ is a discrete set, with its elements being separated by a strictly positive distance from all others, except for $0$. Every nonzero element of $\sigma(L)$ is an eigenvalue of $L$ with finite multiplicity.
\end{theorem}

We introduce the Arzel\`a-Ascoli theorem, which is a fundamental result in mathematical analysis. This is used to prove compactness of $T_\pi$.

\begin{theorem}[Arzel\`a-Ascoli] \label{thm: arzela-ascoli}
    Let $F$ be a set of real-valued continuous functions defined on a compact metric space $(X, d)$.
    Suppose $F$ is uniformly bounded, i.e., there exists an $M \in \mathbb{R}$ such that $\sup_{x \in X} |f(x)| \leq M$ for all $f \in F$.
    If this set is uniformly equicontinuous, i.e., for every $\epsilon > 0$, there exists a $\delta > 0$ (depending on $\epsilon$ but not on $x, y$) such that $|f(x) - f(y)| < \epsilon$ whenever $d(x,y) < \delta$ for all $f \in F$, then $F$ is relatively compact.
\end{theorem}

\begin{table}
    \centering
    \caption{Practical Implications of the Theorems for EigenSafe}
    \label{tab:theory_guide}
    \begin{tabular}{m{3.5cm} m{4.5cm}} \toprule
        \multicolumn{1}{c}{\textbf{Mathematical Guarantee}} & \multicolumn{1}{c}{\textbf{Practical Implication}} \\
        \midrule
        \autoref{thm: compactness of T}: The operator $T_{\pi}$ is compact. & 
        Enables the subsequent theoretical analyses by guaranteeing that $T_\pi$ has similar properties to finite matrices. \\
        \midrule
        \autoref{thm: dominant eigenfunction positivity}: 
        The dominant eigenfunction $\phi_{\pi}$ is strictly positive everywhere in the safe set. & 
        Supports that the dominant eigenfunction plays a role of a calibrated safety critic, (the safety probability is always positive). \\
        \midrule
        \autoref{thm: uniqueness}: 
        The dominant eigenpair $(\gamma_{\pi}, \phi_{\pi})$ is unique up to scaling. & 
        Supports that the learning algorithm has a stable, unique target to converge to, avoiding mode collapse. \\
        \midrule
        \autoref{thm: spectral gap}:
        A strictly positive gap exists between $\gamma_{\pi}$ and the rest of the spectrum $|\lambda| < \gamma_{\pi}$. &
        Ensures that the non-dominant eigenmodes of $T_\pi$ decay faster than the dominant mode, explains the convergence of power iteration. \\
        \midrule
        \autoref{thm: probability approximation}:
        The true safety probability $Z_{\pi}(t,x)$ decays asymptotically as $c \cdot \phi_{\pi}(x) \cdot \gamma_{\pi}^t$. &
        Justifies the use of the dominant eigenvalue $\gamma_{\pi}$ and the dominant eigenfunction $\phi_\pi$ as global and local safety scores, respectively. \\
        \midrule
        \autoref{thm: non-dominant}:
        All real, non-dominant eigenfunctions must take negative values. & 
        Justifies the auxiliary positivity loss $\mathcal{J}_{+}$ introduced in \autoref{eq: positivity loss}. This loss penalizes converging to any non-dominant mode. \\
        \midrule
        \autoref{thm: global-local equivalence}:
        The global spectral constraint $\gamma_{\pi} \ge \gamma_0$ is equivalent to a local condition on $T_{\pi}$. & 
        Enables the implementation of the global safety constraint using only local transitions sampled from the replay buffer. \\
        \bottomrule
    \end{tabular}
\end{table}

\subsection{Compactness of $T_\pi$}

We show that $T_\pi$ is a compact operator having the favorable properties discussed above. This requires the assumption that the closed-loop transition density function $p_\pi(\cdot|\cdot)$ is continuous and that the safe set $C$ is a compact subset of a Euclidean space.

\begin{assumption}[Compactness of $C$] \label{assumption: C compactness}
    The set $C$ is a compact subset of a Euclidean space with a non-empty interior.
\end{assumption}

\begin{assumption}[Continuity of Transition Density] \label{assumption: kernel continuity}
    The closed-loop transition probability density $p_\pi(y|x)$ is a continuous function of $(x,y) \in C \times C$.
\end{assumption}

\begin{theorem}[Compactness of $T_\pi$] \label{thm: compactness of T}
    Under Assumptions~\ref{assumption: C compactness} and \ref{assumption: kernel continuity}, $T_\pi$ is a compact operator.
\end{theorem}
\begin{proof}
    Let $\mathcal{B} \coloneqq \{\beta \in D_T \;|\; \norm{\beta}_\infty \leq 1\}$ be the unit ball of the domain $D_T$ centered at the origin. We will show that the image $T_\pi \mathcal{B}$ is relatively compact. According to \autoref{thm: arzela-ascoli}, it suffices to show that $T_\pi \mathcal{B}$ is uniformly bounded and uniformly equicontinuous.

    The uniform boundedness of $T_\pi\mathcal{B}$ comes directly from the non-expansive nature of $T_\pi$.
    Next, since $p_\pi(y|x)$ is continuous on a compact set $C\times C$, it is uniformly continuous. That is, for any $\epsilon > 0$, there exists a $\delta > 0$ such that 
    $|x_1 - x_2| < \delta$ implies $|p_\pi(y|x_1) - p_\pi(y|x_2)| < \epsilon / \operatorname{Vol}(C)$ for any $y \in C$. Therefore, for any $\beta \in \mathcal{B}$ and $x_1, x_2 \in C$ with $|x_1 - x_2| < \delta$,
    \begin{equation}
    \begin{aligned}
        |T_\pi \beta(x_1) - T_\pi \beta(x_2)| &= \left|\int_C \beta(y) \left( p_\pi(y|x_1) - p_\pi(y|x_2) \right) dy\right| \\
        &\leq \int_C |\beta(y)| \cdot \left|p_\pi(y|x_1) - p_\pi(y|x_2)\right|dy \\
        &\leq \norm{\beta}_\infty \cdot \frac{\epsilon}{\operatorname{Vol}(C)} \cdot \operatorname{Vol}(C) \leq \epsilon,
    \end{aligned}
    \end{equation}
    proving uniform equicontinuity.
    Here, $\operatorname{Vol}(C) \coloneqq \int_C dy$.
\end{proof}

As a corollary, we also find that every eigenfunction of $T_\pi$ corresponding to a nonzero eigenvalue is continuous on $C$.
\begin{corollary}[Continuity of Eigenfunction] \label{thm: eigenfunction continuity}
    Under Assumptions~\ref{assumption: C compactness} and \ref{assumption: kernel continuity}, every eigenfunction of $T_\pi$ for a nonzero eigenvalue is continuous on $C$.
\end{corollary}
\begin{proof}
    Since $p_\pi(y|x)$ is jointly continuous in $x$ and $y$, the operator $T_\pi$ defined as $T_\pi \beta(x) = \int_C \beta (y) p_\pi (y|x) dy$ maps \textit{any} bounded measurable function to a continuous function. 
    This can be shown using the same technique as in \autoref{thm: compactness of T}.
    Since every eigenfunction for a nonzero eigenvalue lies within the image of $T_\pi$, it must be continuous.
\end{proof}

\subsection{Nonnegativity of the Dominant Eigenpair}

In this subsection, we see that the dominant eigenpair of $T_\pi$ is nonnegative real.
Recall that $T_\pi$ is a positive operator. Mathematically, we define the positive cone $\mathcal{K} \subset D_T$ as
\begin{equation}
    \mathcal{K} \coloneqq \{\beta \in D_T \;|\; \beta(x) \geq 0,\; \forall x \in S\}.
\end{equation}
This $\mathcal{K}$ is a \textit{total cone}, meaning that $\mathcal{K} \cap (-\mathcal{K}) = \{0\}$ and $\mathcal{K} + (-\mathcal{K}) = D_T$.
The operator $T_\pi$ leaves this cone invariant ($T_\pi \mathcal{K} \subseteq \mathcal{K}$) and is therefore a positive operator. 

\begin{theorem}[Krein-Rutman \cite{krein_rutman}] \label{thm: krein-rutman}
    Let $\mathcal{D}$ be a Banach space and $\mathcal{K} \subset \mathcal{D}$ be a total cone.
    Let $T:\mathcal{D}\rightarrow \mathcal{D}$ be a nonzero compact operator such that
    \begin{equation}
        T\mathcal{K} \subseteq \mathcal{K},
    \end{equation}
    and that its spectral radius $\rho(T)$ is strictly positive. Then, $\rho(T)$ is an eigenvalue of $T$ with the corresponding eigenvector $v$ being positive, i.e., $v \in \mathcal{K} \setminus \{0\}$.
\end{theorem}

Since $T_\pi$ is compact as already discussed, the Krein-Rutman theorem guarantees that the spectral radius $\gamma_\pi = \rho(T_\pi)$ is an eigenvalue, whose corresponding eigenfunction $\phi_\pi$ is nonnegative everywhere on $C$.

\subsection{Uniqueness of the Dominant Eigenfunction}

Now we show that the dominant eigenfunction $\phi_\pi$ is unique up to scale. This relies on the system being \textit{topologically irreducible} on the safe set $C$. 
Informally, this means that under $\pi$, the system can \textit{reach anywhere} in $C$ before arriving at $K$, with nonzero probability.

\begin{assumption}[Topological Irreducibility] \label{assumption: topological irreducibility}
    For any $x \in C$ and any non-empty open\footnote{We consider the subspace topology on $C$.} subset $U$ of $C$, there exists a time horizon $n$ such that
    \begin{equation}
        \mathbb{P}_\pi \left[s_n \in U \middle| s_0 = x\right] > 0.
    \end{equation}
\end{assumption}

One property that we can derive under \autoref{assumption: topological irreducibility} is that $\phi_\pi$ has strictly positive values in $C$.
\begin{theorem} \label{thm: eigenfunction positivity}
    Under Assumptions~\ref{assumption: C compactness}, \ref{assumption: kernel continuity}, and \ref{assumption: topological irreducibility}, any nonnegative real eigenfunction of $T_\pi$ is strictly positive.
\end{theorem}
\begin{proof}
    Let the eigenpair be $(\lambda, \phi)$.
    Since $T_\pi \phi = \lambda \phi$ and $T_\pi$ is a positive operator, $\lambda \geq 0$.
    Firstly, $\phi$ is a continuous function (\autoref{thm: eigenfunction continuity}). Consider the set $U$ defined as 
    \begin{equation}
        U = \{x\in C \;|\; \phi(x) > 0\},
    \end{equation}
    which is a non-empty open subset of $C$.
    Let $x$ be an arbitrary point in $C$.
    \autoref{assumption: topological irreducibility} says that there is a $n \in \mathbb{N}$ such that
    \begin{equation} \label{eq: select n}
        \mathbb{P}_\pi [s_n \in U | s_0 = x] > 0.
    \end{equation}
    Let $p^n_\pi(y|x)$ be the $n$-step transition probability density function under the policy $\pi$, i.e., $T_\pi^n \beta(x) = \int_C \beta(y) p^n_\pi(y|x) dy$. Since $T_\pi^n$ should have $(\lambda^n, \phi)$ as an eigenpair,
    \begin{equation}
    \begin{aligned}
        \lambda^n \phi(x) &= T_\pi^n \phi(x) \\
        & = \mathbb{E}_\pi [\phi(s_n) | s_0 = x] \\
        &= \int_C \phi (y) p^n_\pi(y|x) dy \\
        &= \underbrace{\int_U \phi (y) p^n_\pi(y|x) dy}_{\text{(A)}} + \underbrace{\int_{C \setminus U}\phi (y) p^n_\pi(y|x) dy}_{\text{(B)}}.
    \end{aligned}
    \end{equation}
    Since \autoref{eq: select n} and $\phi (y) > 0$ for all $y \in U$, $\text{(A)}$ is strictly positive, and since $\phi(y) \geq 0$ everywhere in $C$, $\text{(B)}$ is nonnegative.
    This makes the right hand side strictly positive, and hence $\lambda^n \phi(x) > 0$. This should hold for every $x \in C$, so $\phi$ is strictly positive everywhere in $C$ and $\lambda > 0$.
\end{proof}

\begin{corollary}[Strict Positivity of the Dominant Eigenfunction] \label{thm: dominant eigenfunction positivity}
    Under Assumptions~\ref{assumption: C compactness}, \ref{assumption: kernel continuity}, and \ref{assumption: topological irreducibility}, the dominant eigenfunction $\phi_\pi$ of $T_\pi$ is strictly positive everywhere in $C$.
\end{corollary}

We can use this result to prove that $\phi_\pi$ is a unique dominant eigenfunction corresponding to the eigenvalue $\gamma_\pi$.

\begin{theorem}[Uniqueness of the Dominant Eigenfunction] \label{thm: uniqueness}
    Under Assumptions~\ref{assumption: C compactness}, \ref{assumption: kernel continuity}, and \ref{assumption: topological irreducibility}, $\gamma_\pi$ is an eigenvalue of $T_\pi$ with multiplicity $1$, i.e., the eigenfunction $\phi_\pi$ is unique up to scale.
\end{theorem}
\begin{proof}
    We show this by contradiction.
    Suppose that, besides $\phi_\pi \in \mathcal{K}$, there is another eigenfunction $\phi \in D_T \setminus \{0\}$ corresponding to the eigenvalue $\gamma_\pi$, which is linearly independent of $\phi_\pi$.
    Since $T_\pi$ is a real operator (it maps real functions to real functions), $\phi$ can be picked to have real values everywhere.
    Consider the set $M \coloneqq \{c \in \mathbb{R} \;|\; c \phi_\pi (x) - \phi(x) \geq 0, \forall x \in C\}$. Since $\phi_\pi(x) > 0$ for all $x \in C$ (\autoref{thm: dominant eigenfunction positivity}), $M$ should have an infimum $\alpha = \inf M$. Let $h \coloneqq \alpha \phi_\pi - \phi$, and with this $\alpha$, there exists an $x_0 \in C$ such that $h(x_0) = 0$ (otherwise, $\alpha$ is not the infimum of $M$).
    
    Since $h$ is a linear combination of eigenfunctions corresponding to the same eigenvalue, $h$ is also an eigenfunction for $\gamma_\pi$ satisfying $T_\pi h = \gamma_\pi h$, which is nonnegative everywhere and $h(x_0) = 0$ for an $x_0 \in C$.
    Moreover, since we have assumed $\phi_\pi$ and $\phi$ are linearly independent, $h$ cannot be identically zero.
    This contradicts \autoref{thm: eigenfunction positivity}, proving that the dominant eigenfunction of $T_\pi$ is unique up to scalar multiplication.
\end{proof}

\subsection{On the Existence of a Nonzero Spectral Gap}

\begin{figure}
    \centering
    \begin{tikzpicture}[
        >=Stealth,
        node distance=2.5cm,
        every state/.style={thick, fill=gray!10},
        auto
    ]
        \node[state] (x1) {$x_1$};
        \node[state] (x2) [right=of x1] {$x_2$};
        \node[state] (K) [below=1.5cm of $(x1)!0.5!(x2)$] {$K$};
        \path[->, thick] 
            (x1) edge [bend left=20] node {0.5} (x2)
            (x1) edge [bend right=20] node [swap] {0.5} (K)
            (x2) edge [bend left=20] node {0.5} (x1)
            (x2) edge [bend left=20] node {0.5} (K)
            (K) edge [loop below] node {1} (K);
    \end{tikzpicture}
    \caption{This Markov chain satisfies \autoref{assumption: topological irreducibility} but not \ref{assumption: mixing}. It has a non-positive eigenvalue whose absolute value equals the spectral radius of $T_\pi$. The numbers on the arrows denote transition probabilities.}
    \label{fig: markov_chain}
\end{figure}

Although we have demonstrated the uniqueness of the eigenfunction corresponding to the eigenvalue $\gamma_\pi$, this does not automatically eliminate the possibility of another eigenvalue, either complex or negative, having the same absolute value as $\gamma_\pi$. For example, consider a discrete-state system with state space $S = \{x_1, x_2, K\}$ and transition probabilities as illustrated in \autoref{fig: markov_chain}. In this scenario, the temporal evolution of $Z_\pi$ is fully described using a $2 \times 2$ matrix as follows.
\begin{equation}
    \begin{bmatrix}
        Z_\pi (t+1, x_1) \\
        Z_\pi (t+1, x_2)
    \end{bmatrix} = \begin{bmatrix}
        0 & 0.5 \\
        0.5 & 0
    \end{bmatrix}
    \begin{bmatrix}
        Z_\pi (t, x_1) \\
        Z_\pi (t, x_2)
    \end{bmatrix}
\end{equation}
It is straightforward to find that this process is topologically irreducible, i.e., it satisfies \autoref{assumption: topological irreducibility}. However, besides the dominant eigenpair $(0.5, [1, 1]^\top)$, there exists another eigenpair $(-0.5, [1, -1]^\top)$ whose eigenvalue has an absolute value of $0.5$, which equals the spectral radius of $T_\pi$.

For a nonzero spectral gap, we need a stricter condition, which we introduce below.
\begin{assumption}[Uniform Topological Mixing] \label{assumption: mixing}
    There exists a \textit{uniform} horizon $n\in \mathbb{N}$ such that for any non-empty open subset $U \subseteq C$ and any state $x \in C$, 
    \begin{equation}
        \mathbb{P}_\pi \left[s_n \in U \; \middle|\; s_0 = x\right] > 0,
    \end{equation}
    with $n$ not depending on the choice of $x$ and $U$.
\end{assumption}
This implies that
\begin{equation}
    \int_U p^n_\pi (y|x) dy > 0
\end{equation}
for all $x$ and $U$, hence $p^n_\pi(y|x) > 0$ for all $x$ and $y$, except for those in a measure-zero subset of $C \times C$.
Note that \autoref{assumption: topological irreducibility} automatically holds if \autoref{assumption: mixing} is satisfied.

\begin{theorem}[Nonzero Spectral Gap] \label{thm: spectral gap}
    Under Assumptions \ref{assumption: C compactness}, \ref{assumption: kernel continuity}, and \ref{assumption: mixing}, $T_\pi$ has a nonzero spectral gap.
\end{theorem}
\begin{proof}
    Through \autoref{thm: spectral theorem}, we know that every nonzero element of $\sigma(T_\pi)$ is an eigenvalue, which is separated from all other eigenvalues by a nonzero distance. Thus, it suffices to show that $\gamma_\pi$ is the only eigenvalue having modulus $\gamma_\pi$.
    
    Let $\phi$ be an eigenfunction of $T_\pi$ for the eigenvalue $\lambda$ with $|\lambda| = \gamma_\pi$.
    Taking the absolute value gives
    \begin{equation} \label{eq: inequality to equality}
    \begin{aligned}
        \gamma_\pi |\phi | (x) &= |\lambda||\phi(x)| = |T_\pi\phi(x)| = \left|\int_C p_\pi (y|x) \phi(y) dy \right| \\
        &\leq \int_C p_\pi(y|x) |\phi(y)| dy = T_\pi |\phi|(x),
    \end{aligned}
    \end{equation}
    where $|\phi| \in D_T$ is defined as $|\phi|(x) = |\phi(x)|$ for all $x\in C$. If there is an $x \in C$ such that \autoref{eq: inequality to equality} is satisfied with the strict inequality ($<$), then this leads to a contradiction to that $\gamma_\pi$ is the spectral radius of $T_\pi$.
    Thus, \autoref{eq: inequality to equality} should be satisfied with the equality ($=$) everywhere on $C$, making $|\phi|$ a dominant eigenfunction.
    Since the dominant eigenfunction is unique up to scale (\autoref{thm: uniqueness}), we can let $\phi(x) = \phi_\pi(x) e^{\mathrm{i}\theta(x)}$ for a real scalar function $\theta:C\rightarrow \mathbb{R}$.
    Now, observe that
    \begin{equation}
    \begin{multlined}
        \left|\int_C p_\pi^n (y|x) \phi(y) dy  \right| = |T_\pi^n \phi (x)| = |\lambda^n \phi (x)| = \gamma_\pi^n \phi_\pi(x) \\
        = T_\pi^n \phi_\pi (x) = T_\pi^n |\phi| (x) = \int_C p_\pi^n (y|x) |\phi (y)| dy.
    \end{multlined}
    \end{equation}
    Since $p_\pi^n(y|x)$ is strictly positive almost everywhere on $C\times C$, for the equality $|\int_C p_\pi^n(y|x) \phi(y) dy| = \int_C p_\pi^n(y|x) |\phi(y)| dy$ to hold, the complex argument (phase) of $\phi(y)$ should be constant almost everywhere. Combining with the continuity of $\phi$ (\autoref{thm: eigenfunction continuity}), we conclude that $\phi = c\cdot \phi_\pi$ for a constant $c \in \mathbb{C}$.
\end{proof}

\subsection{On the Asymptotic Approximation of $Z_\pi$}

In this subsection, we show the validity of the probability approximation
\begin{equation*}
    Z_\pi(t,x) = c\cdot \phi_\pi(x) \cdot \gamma_\pi^t + o(\gamma_\pi^t),
\end{equation*}
which was introduced in \autoref{eq: prob approximation} (and in a similar form in \autoref{eq: H approximation}).

\begin{theorem} \label{thm: probability approximation}
    Under Assumptions~\ref{assumption: C compactness}, \ref{assumption: kernel continuity}, and \ref{assumption: mixing}, there exists a constant $c \in \mathbb{R}$ such that
    \begin{equation}
        \lim_{t \rightarrow \infty} \sup_{x \in C} \left|\frac{Z_\pi(t,x)}{\gamma_\pi^t} - c \cdot \phi_\pi(x)\right| = 0.
    \end{equation}
\end{theorem}
\begin{proof}
    Since the operator $T_\pi$ is compact, its eigenvalues accumulate only at $0$.
    Moreover, the irreducibility condition says that the multiplicity of the dominant eigenvalue is always $1$ (\autoref{thm: uniqueness}).
    Consider the Riesz projector
    \begin{equation}
        P \coloneqq \frac{1}{2\pi \mathrm{i}}\oint_{\Gamma} \left(z\cdot \operatorname{id} - T_\pi\right)^{-1} dz,
    \end{equation}
    where $\Gamma$ is a positively-oriented contour with its interior containing $\gamma_\pi$ but no other eigenvalues of $T_\pi$. Constructing this contour is possible because $T_\pi$ is a compact operator and therefore has isolated nonzero eigenvalues. This $P$ is the \textit{projection} map onto the one-dimensional eigenspace for $\gamma_\pi$.
    Thus, the operator $T_\pi$ can be decomposed as
    \begin{equation}
        T_\pi = \gamma_\pi P + N,
    \end{equation}
    with $N = T_\pi \circ (\operatorname{id} - P)$, such that $PT_\pi = T_\pi P = \gamma_\pi P$, $PN = NP = 0$, and $N$ is also a compact operator with $\sigma(N) = \sigma(T_\pi) \setminus \{\gamma_\pi\}$.
    We can directly apply this to the relation $Z_\pi(t,\cdot) = T_\pi^t 1_C(\cdot)$ to get
    \begin{equation}
        Z_\pi(t,\cdot) = T_\pi^t 1_C = \gamma_\pi^t P 1_C + N^t 1_C,
    \end{equation}
    and since $P$ is the projection onto the one-dimensional eigenspace spanned by $\phi_\pi$, $P1_C = c\cdot \phi_\pi$ with an appropriate constant $c$. 
    By Gelfand's formula, $\lim_{t \rightarrow \infty} \norm{N^t}^{1/t} = \rho(N)$. Since the spectral radius of $N$ must be strictly smaller than $\gamma_\pi$ (\autoref{thm: spectral gap}), the residual term $N^t 1_C$ decays faster than $\gamma_\pi^t$, i.e., $N^t 1_C = o(\gamma_\pi^t)$.
\end{proof}

\subsection{On the Utility of the Additional Loss Term $\mathcal{J}_+$}

Now, we discuss the utility of the additional loss term
\begin{equation*}
    \mathcal{J}_+[\psi] = \frac{W_+}{|\mathcal{D}|}\sum_{(x,u,\cdot) \in \mathcal{D}} \operatorname{ReLU}\left(-\psi(x,u)\right)
\end{equation*}
introduced in \autoref{eq: positivity loss}.
This loss is optionally employed to accelerate the convergence of the eigenpair learning process in the early stage of training.
The rationale is based on the fact that every non-dominant eigenfunction is either complex-valued or sign-changing (taking both positive and negative values on $C$), which we show below.
Since the neural network function approximator is restricted to real values, the loss $\mathcal{J}_+$ eliminates the possibility of the learning process mis-converging to any of the non-dominant modes and thus effectively biases the optimization towards the unique (\autoref{thm: uniqueness}) dominant eigenpair.

\begin{theorem} \label{thm: non-dominant}
    Under Assumptions~\ref{assumption: C compactness}, \ref{assumption: kernel continuity}, and \ref{assumption: topological irreducibility}, every real non-dominant eigenfunction of $T_\pi$ takes both positive and negative values.
\end{theorem}
\begin{proof}
    We prove this by contradiction. Assume $\phi$ is a real, nonnegative, non-dominant eigenfunction of $T_\pi$, corresponding to the eigenvalue $\lambda \neq \gamma_\pi$. Following a similar procedure as in \autoref{thm: eigenfunction positivity}, we easily find that $\lambda$ is positive real and $\lambda < \gamma_\pi$, and that $\phi$ is strictly positive everywhere in $C$. Since $\phi$ and $\phi_\pi$ are both positive everywhere, we can apply an appropriate scaling to each of them, such that $\min_{x \in C} \phi(x) \geq 1$ and $\max_{x \in C} \phi_\pi (x) = 1$.
    With this, we have the relation
    \begin{equation}
        \phi(x) \geq 1 \geq \phi_\pi(x), \quad \forall x \in C.
    \end{equation}
    Since $T_\pi$ is a positive operator, $T_\pi^n (\phi - \phi_\pi) \in \mathcal{K}$ for any $n \in \mathbb{N}$, i.e.,
    \begin{equation} \label{eq: positive_negative}
        \lambda^n \phi(x) = T_\pi^n \phi(x) \geq T_\pi^n \phi_\pi(x) = \gamma_\pi^n \phi_\pi(x), \; \forall n \in \mathbb{N}, x \in C.
    \end{equation}
    On the other hand, for any $x_0 \in C$, pick a sufficiently large $n$ satisfying
    \begin{equation} \label{eq: logarithm n}
        n > \frac{\log \left(\phi(x_0)/ \phi_\pi(x_0)\right)}{\log (\gamma_\pi / \lambda)},
    \end{equation}
    such that $\lambda^n \phi(x_0) < \gamma_\pi^n$. Note that the denominator in \autoref{eq: logarithm n} is strictly positive because $\lambda < \gamma_\pi$.
    With this $n$,
    \begin{equation}
        \lambda^n \phi(x_0) < \gamma_\pi^n \leq \gamma_\pi^n \phi_\pi(x_0).
    \end{equation}
    This contradicts \autoref{eq: positive_negative}, proving such eigenpair $(\lambda, \phi)$ cannot exist.
\end{proof}

\subsection{Equivalence of Global and Local Safe RL Constraints}

In the safe RL formulation using EigenSafe (\autoref{sec: rl}), we transformed the global eigenvalue constraint 
\begin{equation}
    \gamma_\pi \geq \gamma_0
\end{equation}
to the local state-action constraint
\begin{equation}
    \underbrace{\mathbb{E}_{x' \sim P(\cdot|x,u), u'\sim \pi(\cdot|x')} \psi_\pi(x', u')}_{{} = A_\pi \psi_\pi(x,u)} \geq \gamma_0 \psi_\pi(x,u).
\end{equation}
In this subsection, we prove that the two constraints are equivalent.

\begin{theorem} \label{thm: global-local equivalence}
    Suppose Assumptions~\ref{assumption: C compactness} and \ref{assumption: kernel continuity} hold. Let $\gamma_0 \geq 0$ be a nonnegative scalar. The following two statements are equivalent:
    \begin{enumerate}[(a)]
        \item There exists a $\beta \in \mathcal{K} \setminus \{0\}$ such that $T_\pi \beta(x) \geq \gamma_0 \beta(x)$ for all $x \in C$.
        \item $\gamma_\pi \geq \gamma_0$.
    \end{enumerate}
\end{theorem}
\begin{proof}
    The implication (b) $\Rightarrow$ (a) follows directly by choosing $\beta = \phi_\pi$. To show (a) $\Rightarrow$ (b), we start by observing that
    \begin{equation}
        T_\pi^n \beta (x) \geq \gamma_0 T_\pi^{n-1} \beta (x) \geq \cdots \geq \gamma_0^n \beta (x),
    \end{equation}
    for all $x \in C$ and $n \in \mathbb{N}$, which holds because $T_\pi$ is a positive operator.
    Since both sides are nonnegative,
    \begin{equation}
        \norm{T_\pi^n \beta}_\infty \geq \gamma_0^n \norm{\beta}_\infty,
    \end{equation}
    and by the definition of the operator norm \autoref{eq: operator norm}, $\norm{T_\pi^n \beta}_\infty \leq \norm{T_\pi^n} \cdot \norm{\beta}_\infty$. Thus, we get
    \begin{equation}
        \norm{T_\pi^n}^{1/n} \geq \gamma_0.
    \end{equation}
    Since this must hold for all $n$, we can directly apply Gelfand's formula to get $\gamma_\pi = \rho(T_\pi) \geq \gamma_0$.
\end{proof}

\subsection{Discussion on the Assumptions}

While the theoretical guarantees of EigenSafe rely on Assumptions~\ref{assumption: C compactness} through \ref{assumption: mixing}, these conditions are mild and standard in the context of real-world robotic systems. We discuss the validity of each assumption below.

\begin{enumerate}

\item \textbf{On \autoref{assumption: C compactness} (Compactness of $C$).}

Physical robotic systems operate within bounded state spaces due to mechanical joint limits, actuator saturation, and finite workspace boundaries. Therefore, the assumption that the safe set $C$ is compact is consistent with the physical constraints of the hardware.

\item \textbf{On \autoref{assumption: kernel continuity} (Continuity of $p_\pi$). }

The dynamics of physical systems, governed by classical mechanics, are generally continuous functions of state and action.
Even in case where the dynamics exhibit discontinuities (such as contact dynamics), the transition density $p_\pi$ typically remains continuous due to the stochasticity inherent in real-world environments. Mathematically, this is because the convolution of the deterministic dynamics with a continuous noise distribution (e.g., Gaussian sensor noise) yields a smooth transition density.

\item \textbf{On Assumptions~\ref{assumption: topological irreducibility} and \ref{assumption: mixing} (Irreducibility and Mixing).}

These assumptions require that the safe set is \textit{connected} through the dynamics and that the robot does not enter rigid, deterministic limit cycles.
In practice, given nonzero noise, the inherent stochasticity (arising from environmental noises or complexity of the system itself) ensures the system can at least locally explore and eventually reach any part of the connected safe region. This prevents the system from becoming \textit{locked} in deterministic cycles.
While the mixing horizon $n$ in \autoref{assumption: mixing} may be large for systems with low noise, the existence of such a horizon is sufficient for the theoretical guarantees to hold.
\end{enumerate}

\clearpage

\onecolumn %%%%%%%%%%%%%%%%%

\section{Safe Reinforcement Learning using EigenSafe: Experimental Details}

\subsection{Description of the Gym Environments} \label{sec: gym description}

\begin{figure*}[ht]
    \centering
    \includegraphics[width=0.8\linewidth]{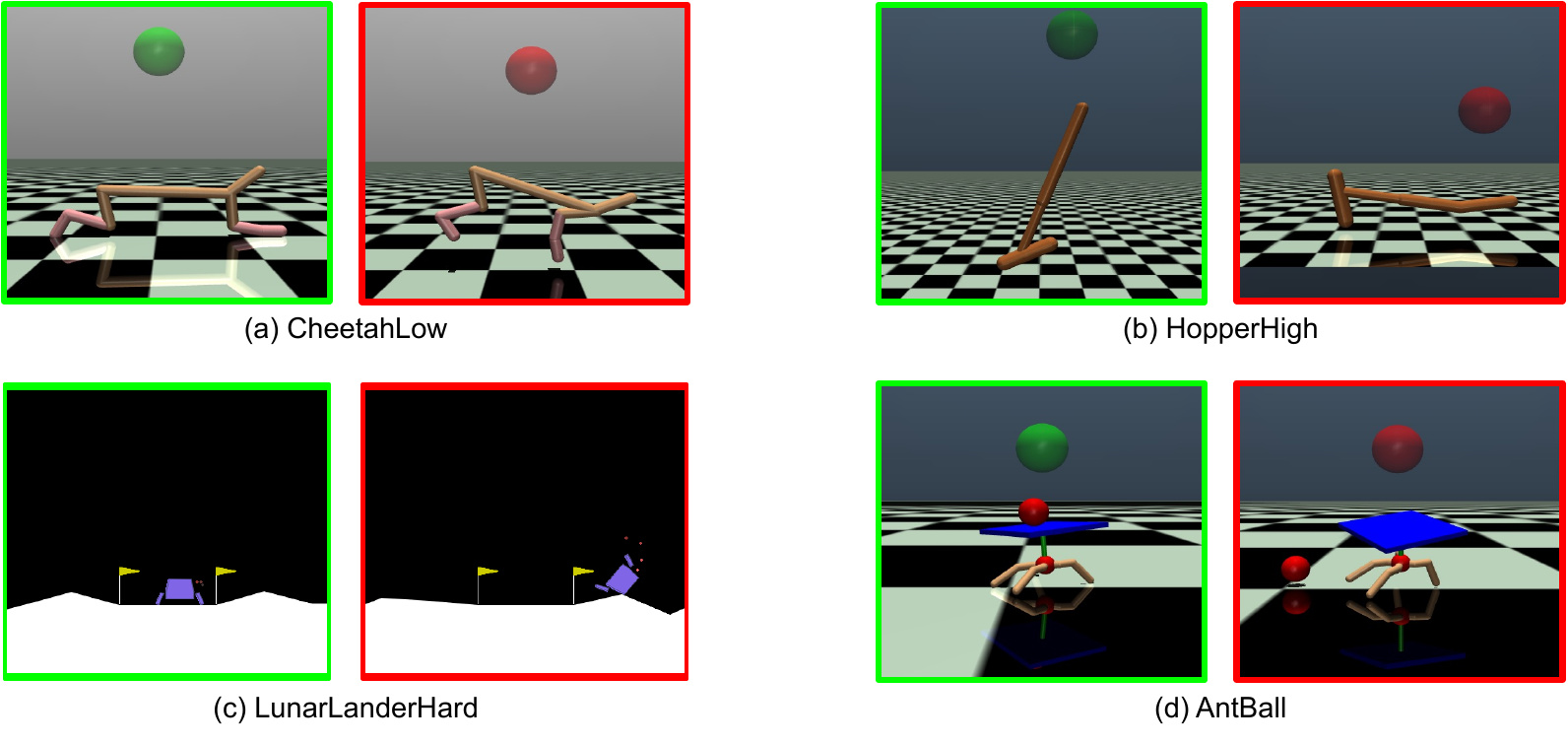}
    \caption{Safe (green-framed) and unsafe (red-framed) states in the Gym environments used in the safe RL experiments.}
    \label{fig:1}
\end{figure*}

In this subsection, we provide more detailed information regarding the Gym \cite{2016openai} environments used in the safe RL demonstration in \autoref{sec: rl} (\autoref{fig: safe rl environments}, \autoref{fig:1}).

\begin{enumerate}[(a)]
\item \texttt{CheetahLow} is adapted from the standard \texttt{HalfCheetah} environment.
Its state consists of proprioceptive joint positions and velocities, while the action corresponds to joint torque commands.
The resulting state and action spaces are continuous, with dimensions of 17 and 6, respectively, governed by the standard MuJoCo simulator dynamics.
Unlike the default reward formulation which includes additional terms such as control costs and survival bonuses, our reward is designed to depend solely on forward velocity, with safety defined as an independent constraint.
This simplification decouples the roles of the reward signal and the safety constraints and enables a clearer comparative analysis of the algorithms.

The agent is rewarded for maintaining a high forward velocity. To prevent the Q function from exploding at high speeds, the reward saturates at a prescribed velocity threshold. Specifically, the reward signal $r(x,u)$ as a function of state $x$ and action $u$ is designed as follows:
\begin{equation}
    r(x, u) = \min\{2v_x(x), 1\},
\end{equation}
where $v_x(x)$ is the forward velocity of the torso. The safety constraint requires the torso to remain below a fixed height threshold. We define the safe set $C$ as
\begin{equation}
C
=
\left\{
x \;\middle|\;
p_z(x) \le -0.2
\right\},
\end{equation}
where $p_z(x)$ is the vertical position of the torso.
The maximum episode length of \texttt{CheetahLow} is 200 timesteps.

\item \texttt{HopperHigh} is adapted from the standard \texttt{Hopper} environment. Its 11-dimensional state comprises the proprioceptive data (joint angles and angular rates), and the 3-dimensional action represents joint torque commands.
Unlike the default reward formulation in the original \texttt{Hopper} environment, we incentivize the agent only for moving forward:
\begin{equation}
    r(x,u) = \min\{v_x(x), 1\}
\end{equation}
where $v_x(x)$ is the forward velocity of the torso.
The safety constraint requires the height of the torso to be above a given threshold:
\begin{equation}
C
=
\left\{
x \;\middle|\;
p_z(x) \ge 0.85
\right\},
\end{equation}
where $p_z(x)$ is the vertical position of the torso. Its maximum episode length is 400 timesteps.

\item \texttt{LunarLanderHard} is a more challenging variant of the standard \texttt{LunarLander} benchmark with diversified initial states.
The initial lander position is sampled uniformly from a bounded region
$p_x \in [8.5, 11.5]$ and $p_y \in [10.0, 11.0]$.
The 8-dimensional state consists of its pose, linear/angular velocities, and two binary variables indicating whether the two legs are in contact with the ground.
The 2-dimensional action corresponds to continuous thrust control commands.
Unlike the default formulation, which includes additional terms such as velocity penalties and orientation constraints,
we use a simplified reward consisting only of a dense shaping term based on the progress toward the landing position,
together with a sparse reward for goal reaching.
Safety-related terms are separated from the original reward formulation and are encoded as constraints.
\begin{equation}
r(x, u)
=
\begin{cases}
30 & \text{(if the lander has landed successfully)}, \\
-10 & \text{(if the lander is outside the frame)}, \\
100\cdot(d(x^-)-d(x)) 
- 0.30\,u_m(u) 
- 0.03\,u_s(u) & \text{(otherwise)},
\end{cases}
\end{equation}
where $x^-$ is the state at the previous time step, $d(x)$ is the Euclidean distance of the lander to the landing target, and $u_m(u)$ and $u_s(u)$ represent the throttle levels of the main and side engines, respectively.
The safety constraint is set as
\begin{equation}
C
=
\left\{
x \;\middle|\;
|v(x)| \le 0.5,
\;
|\theta(x)| \le \pi/4,
\;
6 \le p_x(x) \le 14
\right\},
\end{equation}
where $v(x)$ is the planar velocity, $\theta(x)$ is the orientation, $p(x)$ is the horizontal position of the lander, respectively.
It has a maximum episode length of 400 timesteps.

\item \texttt{AntBall} is a customized version of the standard \texttt{Ant} benchmark, with a square plate rigidly attached to the top of the torso and a ball placed above it. The initial position of the ball is sampled from a uniform distribution over a bounded region $p_x \in [-0.4, 0.4]$ and $p_y \in [-0.4, 0.4]$, where $p_x$ and $p_y$ denote the horizontal coordinates of the ball. The state is 30-dimensional, comprising the joint angles and velocities, together with the 3-dimensional relative position of the ball with respect to the torso. The 8-dimensional action corresponds to joint torque commands.
Unlike the standard \texttt{Ant} environment, the agent is incentivized only for its forward velocity, and the safety constraint is defined with respect to the vertical position of the ball, which becomes naturally violated when the robot drops it:
\begin{equation}
    r(x,u) = \min\{2v_x(x), 1\}, \quad \quad C = \left\{x \;\middle|\; p_z(x) \geq 1 \right\},
\end{equation}
where $v_x(x)$ is the forward velocity of the torso, $p_z(x)$ is the height of the ball.
The maximum episode length of the \texttt{AntBall} environment is 400 timesteps.

\end{enumerate}

\clearpage

\subsection{Learning Curves}

\begin{figure*}[ht]
    \centering
    \includegraphics[width=0.93\linewidth]{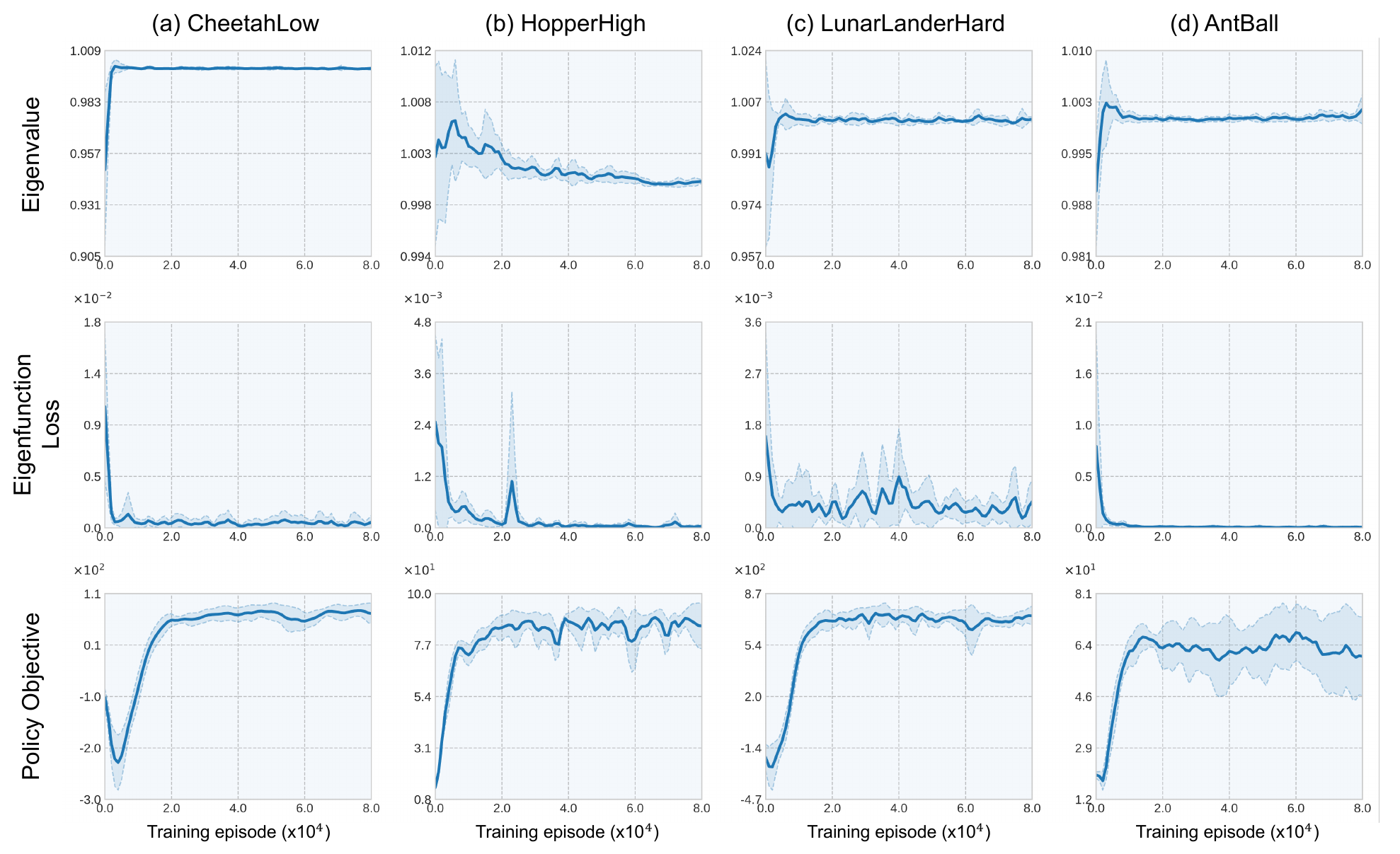}
    \caption{Learning curves of EigenSafe in Gym experiments.}
    \label{fig:2}
\end{figure*}

\begin{figure*}[ht]
    \centering
    \includegraphics[width=0.92\linewidth]{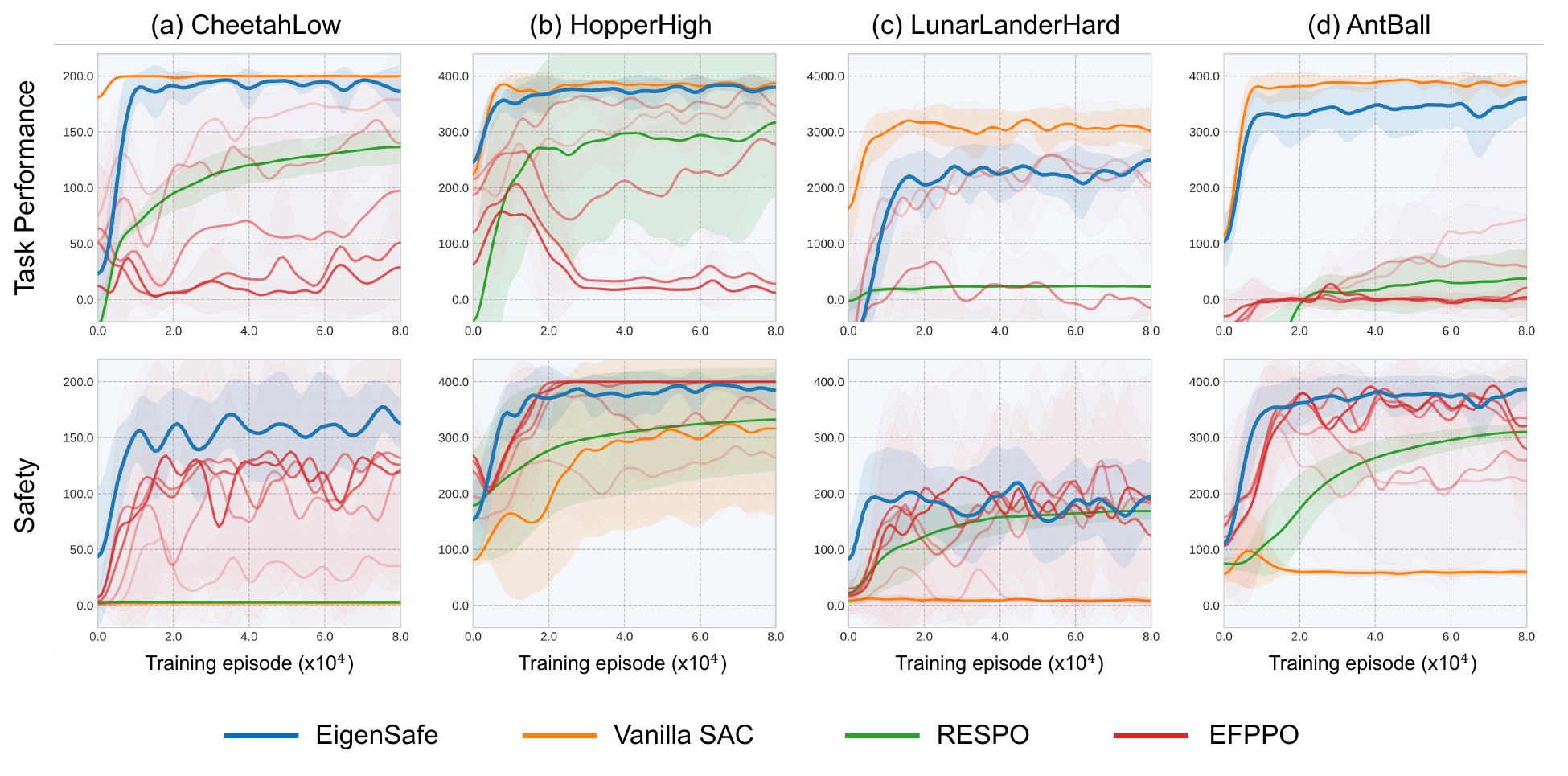}
    \vspace{-1em}
    \hspace{-0.8em} 
    \caption{Learning curves comparing baseline methods in cumulative reward and time to first safety failure.
    For EFPPO, since the policies depend on the initial \textit{safety budget}, we plot multiple curves for each. Thicker curves indicate that the evaluation starts with a higher budget.
    }
    \label{fig:3}
\end{figure*}

\autoref{fig:2} illustrates the learning curves for the eigenfunction loss ($\mathcal{J}_\text{eig}$ as in \autoref{eq: eig loss} with $\mathcal{J}_+$ as in \autoref{eq: positivity loss}) and the policy objective ($\hat{\mathcal{J}}_\text{RL}$ as in \autoref{eq: rl loss augmented}) across each environment, alongside the evolution of the eigenvalue estimate during training. In all tested environments, the learning objectives converged stably, independent of the random network initialization. The eigenvalue estimates gradually increased towards $1$ in all cases, indicating that the learned policy became progressively safer.

In \autoref{fig:3}, we compare EigenSafe against baseline methods in terms of how task performance (which we measure by reward accumulation) and safety (which is measured using time steps until first safety violation) improve over the course of training. The evaluation metrics remain the same as in \autoref{fig: rl performance}.
The apparent trends demonstrate that using EigenSafe as a safety critic achieves a better safety-performance tradeoff. As mentioned in \autoref{sec: rl results}, we attribute this to the eigenfunction critic's ability to assess safety more accurately than reachability-based methods, mitigating unnecessary conservatism.

\begin{table}[h]
\centering
\caption{Hyperparameters for Safe RL with EigenSafe.}

\label{table:3}
\begin{tabular}{cc}
\toprule
\textbf{Parameter} & \textbf{Value} \\
\midrule

Learning rate & \(5 \times 10^{-4}\) \\
Discount factor for SAC Q function $\gamma_\text{RL}$ & \( 0.99 \) \\
Optimizer & Adam \cite{kingma2014adam} \\

\# of episodes per epoch & 10 \\
\# of gradient steps per epoch & 64 \\

Replay buffer size & \(5 \times 10^4\) \\
Minibatch size & 512\\

Target smoothing coefficient & \( 0.995 \) \\
Entropy coefficient & Auto-tuned \cite{haarnoja2018soft} \\

Activation function & ReLU \\

\# of hidden layers & 2 \\
\# of hidden units per layer & 512 \\

Eigenfunction loss weight $W_\text{eig}$ & 1.0 \\ 
Normalization loss weight $W_{n}$ & 1.0 \\
Positivity loss weight $W_\text{+}$ & 1.0 \\

Target eigenvalue $\gamma_0$ & 1.0 \\

Lagrange multiplier relaxation constant $\epsilon$ & \(1 \times 10^{-3}\) \\

\bottomrule
\end{tabular}
\end{table}

\subsection{Implementation Details (Proposed Method)}

The hyperparameter values used in the implementation of the proposed method in the safe RL demonstration (\autoref{sec: rl}) are summarized in \autoref{table:3}.

Theoretically, the Lagrange multiplier $\Lambda$ in \autoref{eq: rl loss augmented} should be formulated as a function of state and action, so that it can account for the varying degrees of safety constraint violation across the domain $S \times A$. However, aligning with the practical implementations of constrained optimization in RL methods,
we find that modeling $\Lambda$ with high expressivity is not critical for the framework. In the implementation, we parametrize $\Lambda$ as a positive scalar rather than a neural network. This single-dimensional $\Lambda$ is updated using the same gradient update rule.
Empirically, we observed that this scalar multiplier grows sufficiently large during training to effectively enforce the safety constraints.

\subsection{Implementation Details (Baseline)}

In this subsection, we provide detailed information regarding the implementation of the baseline methods used for comparison study. 
Since the baselines were not originally developed for exactly the same setting considered in this work, we made appropriate efforts to ensure fairness, e.g., by modifying the original algorithms only minimally and only when necessary, and by using the same hyperparameter values across all baselines.
The problem setting of EFPPO \cite{so2023solving} is not completely identical to the current framework which optimizes a reward-maximizing policy subject to the safety constraints. Instead, EFPPO seeks a policy that minimizes the infinite-horizon policy value function
\begin{equation}
    V^{l,\pi}(x) = \sum_{t=0}^\infty 
    \gamma_\text{RL}^t 
    l(s_t)
\end{equation}
which plays a role as a Lyapunov function,
subject to the safety constraint $s_t \in C$ for all $t \geq 0$,
where $s_t$ is the state trajectory along the closed-loop dynamics starting at $s_0 = x$, and $l$ is a nonnegative cost function. To ensure nonnegativity of $l$, we appropriately scale and bias the reward term so that EFPPO naturally seeks a policy that maximizes the expected return.

\newpage

\section{Safety-Filtered Inference for Imitation Learning using EigenSafe: Experimental Details}

\begin{table}[h]
\centering
\caption{Hyperparameters for Safe IL with EigenSafe.}

\label{table:4}
\begin{tabular}{cc}
\toprule
\textbf{Parameter} & \textbf{Value} \\
\midrule

Learning rate & \(1 \times 10^{-4}\) \\

Optimizer & Adam \cite{kingma2014adam} \\

Minibatch size & 32\\

Activation function & ELU \\

\# of hidden layers & 2 \\
\# of hidden units per layer & 512 \\

\# of LayerNorm layers & 2 \\

Eigenfunction loss weight $W_\text{eig}$ & 1.0 \\ 
Normalization loss weight $W_{n}$ & 1.0 \\
Positivity loss weight $W_\text{+}$ & 1.0 \\

\bottomrule
\end{tabular}
\end{table}

\begin{figure*}[h]
    \centering
    \includegraphics[width=0.7\linewidth]{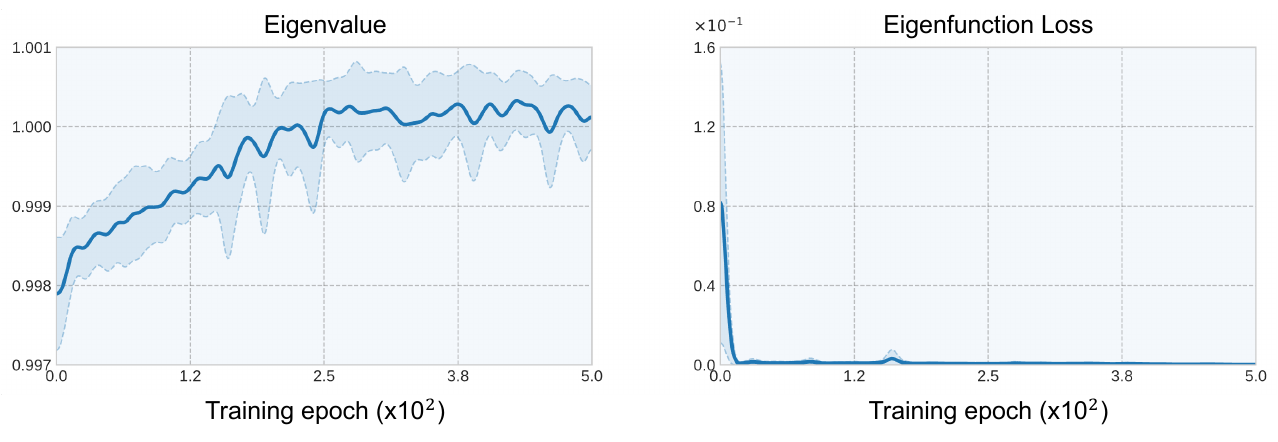}
    \vspace{-1em}
    \caption{Learning curves of EigenSafe in the IL for hardware experiment.}
    \label{fig:5}
\end{figure*}

Finally, we provide additional details on the hardware (UR3) experiment that are not explicitly covered in \autoref{sec: il}.

The two cameras capture $3 \times 224 \times 224$ RGB images at 30 Hz. Images from both cameras are encoded using a frozen, pretrained DINOv2 ViT-S image encoder \cite{oquab2023dinov2}, producing $14 \times 14$ patch tokens, each with dimension 384. We apply mean pooling over the patch dimension, resulting in a single 384-dimensional visual embedding per image. 
The proprioceptive state of the UR3 robotic arm consists of a 7-dimensional end-effector pose and a 1-dimensional gripper state. The visual embeddings from the two cameras are concatenated with the proprioceptive state vector to form the 776-dimensional state which is used as the input to the neural networks.

In the data collection phase, it is crucial to ensure that the dataset captures a diverse range of likely failure modes, in order for the safety-filtering scheme to operate robustly.
Thus, we categorize the failure modes into the following four categories:
\begin{enumerate}
    \item The gripper makes unintended contact with the ground.
    \item Almonds are spilled due to a precarious grasp during the lifting phase.
    \item Almonds are dropped while being carried towards the pan.
    \item Almonds fall outside the pan during the pouring phase.
\end{enumerate}
Although not every failure instance falls strictly into a single category, the dataset used in the experiment consisting of 203 failure demonstrations ensures that each type is represented by at least 50 trajectories.

The learning hyperparameters are summarized in \autoref{table:4}. 
We report the learning curves in \autoref{fig:5}, which exhibit a convergence pattern similar to that observed in the online experiments.

\end{document}